\documentclass[11pt,letterpaper,twoside,reqno,nosumlimits]{amsart}

\usepackage{etoolbox}

\patchcmd{\section}{\scshape}{\bfseries}{}{}
\makeatletter
\renewcommand{\@secnumfont}{\bfseries}
\makeatother

\usepackage[dvipsnames]{xcolor}
\patchcmd{\section}{\normalfont}{\normalfont\color{MidnightBlue}}{}{}
\patchcmd{\subsection}{\normalfont}{\normalfont\color{MidnightBlue}}{}{}

\makeatletter
\def\subsubsection{\@startsection{subsubsection}{3}%
\z@{.5\linespacing\@plus.7\linespacing}{-.5em}%
{\normalfont\bfseries}}
\makeatother

\usepackage{circuitikz}

\usepackage{fancyhdr}
\usepackage{amsmath,amsfonts,amsbsy,amsgen,amscd,mathrsfs,amssymb,amsthm,mathtools,tensor}

\usepackage[a4paper]{geometry}
\usepackage{tikz}
\usepackage{overpic}

\usepackage{stmaryrd}

\usepackage{amsopn}

\DeclareMathOperator*{\minimize}{{\rm minimize}}

\newcommand{\st}{{\rm\,s.t.}}

\usepackage{caption}
\usepackage{graphicx}
\usepackage{color}
\usepackage[bf,SL,BF]{subfigure}
\usepackage{url}
\usepackage{epstopdf}
\usepackage{pdfsync}
\usepackage[colorlinks]{hyperref}
\usepackage[all]{xypic}
\usepackage{comment}
\usepackage{mathabx}
\usepackage{upgreek}

\usepackage{mathrsfs}
\usepackage{yfonts}

\hypersetup{
    linkcolor=blue,
}

\newlength{\fixboxwidth}
\usepackage{epsfig,amsbsy,graphicx,multirow}

\usepackage{ algorithm, algorithmic}
\renewcommand{\algorithmiccomment}[1]{\bgroup\hfill//~#1\egroup}

\usepackage[all,cmtip]{xy}

\usepackage{accents}

 \numberwithin{equation}{section}

\def\V{\mathcal{V}}

\def\R{\mathbb{R}}

\def\cN{\mathcal{N}}

\def\cE{\mathcal{E}}

\def\Y{{\bf\mathcal{Y}}}
\def\Z{\mathcal{Z}}
\def\E{\mathbb{E}}

\def\cR{\mathcal{R}}
\def\B{\mathcal{B}}

\def\X{{\bf\mathcal{X}}}
\def\F{\mathcal{F}}
\def\G{\mathcal{G}}

\def\L{\mathcal{L}}

\def\wK{{\Gamma}}

\def\H{\mathcal{H}}

\def\N{\mathcal{N}}

\def\restrict#1{\raise-.5ex\hbox{\ensuremath|}_{#1}}

\def\<{\big\langle}
\def\>{\big\rangle}

\def\det{\operatorname{det}}

\def\dim{{\operatorname{dim}}}

\def\ba{{\bf a}}

\def\bt{{\bf t}}

\definecolor{red}{rgb}{0.9, 0, 0}

\definecolor{green}{rgb}{0.0, 1.0, 0.0}

\newtheorem{Theorem}{Theorem}[section]

\newtheorem{Remark}[Theorem]{Remark}

\newtheorem{Problem}{Problem}

\makeatletter
\newcommand{\oset}[3][0ex]{%
  \mathrel{\mathop{#3}\limits^{
    \vbox to#1{\kern-2\ex@
    \hbox{$\scriptstyle#2$}\vss}}}}
\makeatother

\makeatletter
\newcommand{\uset}[3][0ex]{%
  \mathrel{\mathop{#3}\limits_{
    \vbox to#1{\kern-2\ex@
    \hbox{$\scriptstyle#2$}\vss}}}}
\makeatother

\usepackage{aurical}
\usepackage[T1]{fontenc}

\usepackage{tikz,pgflibraryshapes}
\usetikzlibrary{shadows}
\usetikzlibrary{arrows}

\begin{document}
%\title{A mechanical regression theory\\  for deep learning}
%\title[Mechanical regression, idea registration]{Mechanical regression, idea registration, and\\  the  elephant in the dark deep learning room}
\title[Computational Graph Completion ]{Computational Graph Completion}
%\title[Mechanical regression, idea registration]{Mechanical regression and idea registration are the continuous limit of artificial neural networks}
%\title{Mechanical regression\\  the  elephant in the dark deep learning room}

\date{\today}

\author{Houman Owhadi}

\thanks{Caltech,  MC 9-94, Pasadena, CA 91125, USA, owhadi@caltech.edu}

\maketitle

\begin{abstract}
We introduce a framework for generating, organizing, and reasoning with computational knowledge.
It is motivated by the observation that most problems in Computational Sciences and Engineering (CSE) can be described as that of completing (from data) a computational graph (or hypergraph) representing dependencies between functions and variables. In that setting nodes represent variables and edges (or hyperedges) represent functions (or functionals).
 Functions and variables may be known, unknown, or random. Data comes in the form of observations of distinct values of a finite number of subsets of the variables of the graph (satisfying its functional dependencies). The underlying problem combines a regression problem  (approximating unknown functions) with a matrix completion problem (recovering unobserved variables in the data). Replacing unknown functions by  Gaussian Processes (GPs) and conditioning on observed data provides a simple but efficient approach to completing such graphs. Since the proposed framework is highly expressive, it has a vast potential application scope.
Since the completion process can be automatized, as one solves $\sqrt{\sqrt{2}+\sqrt{3}}$ on a pocket calculator without thinking about it, one could, with the proposed framework, solve a complex CSE problem by drawing a diagram. Compared to traditional regression/kriging, the proposed framework can be used to recover unknown functions with much scarcer data by exploiting interdependencies between multiple functions and variables.
The Computational Graph Completion (CGC) problem addressed by the proposed framework could therefore also be interpreted as a generalization of that of solving linear systems of equations to that of approximating unknown variables and functions with noisy, incomplete, and nonlinear dependencies. Numerous examples illustrate the flexibility, scope, efficacy, and robustness of the CGC framework and show how it can be used as a pathway to identifying simple solutions to classical CSE problems. These examples include the seamless CGC representation of known methods (for solving/learning PDEs, surrogate/multiscale  modeling, mode decomposition,  deep learning) and the discovery of new ones (digital twin modeling, dimension reduction).
\end{abstract}

\section{Introduction}

\subsection{Motivations}
The complexity of a problem is relative to our ability to decompose it by reasoning through simple concepts encapsulating and abstracting away its intricacies and technicalities. For example, (1) computing $\sqrt{\sqrt{2}+\sqrt{3}}$ is a complex task if one tries to solve it by hand and a simple one if one uses a pocket calculator, (2) calculating a mortgage amortization schedule is a complex task if done by hand and a simple one if done with an Excel sheet,  (3) designing an engine valve is a complex task if done by hand and a simple one if done with AutoCAD. Why? Because the technicalities and intricacies have been abstracted away, and the user is only left with the task of manipulating simple/high-level concepts. Many problems in Computational Sciences and Engineering (CSE) are seen as complex because they involve intricate interplays between multiple physics, multiple scales, nonlinear effects, missing/partial information, and limited resources (in human time and expertise).
Can the complexities of most of these problems be abstracted away? Can the underlying problems be reduced to that of manipulating simple, high-level concepts and relations?
An affirmative answer to these questions would lead to a paradigm shift in which most CSE problems are solved through high-level operations without thinking about them.
The purpose of this manuscript is to introduce a rigorous and unified framework reducing CSE problems to the completion of computational graphs representing dependencies between  (known, unknown or random) functions and variables.
 Replacing unknown functions with Gaussian processes and conditioning on available data/measurements provide a simple but powerful tool for solving Computational Graph Completion (CGC) problems when the underlying kernels are also learned from data \cite{owhadi2019kernel} or programmed \cite{owhadi2019kernelmd}  for the task at hand.
 Since this  completion process can be  automatized, as
 one solves $\sqrt{\sqrt{2}+\sqrt{3}}$ on a pocket calculator without thinking about it, one could, with the proposed framework, solve a complex CSE problem by drawing a diagram (as illustrated below). This would accelerate  innovation and decision making by enabling the real-time development and testing of models, ideas and scenarios.
	\begin{center}
			\includegraphics[width= \textwidth]{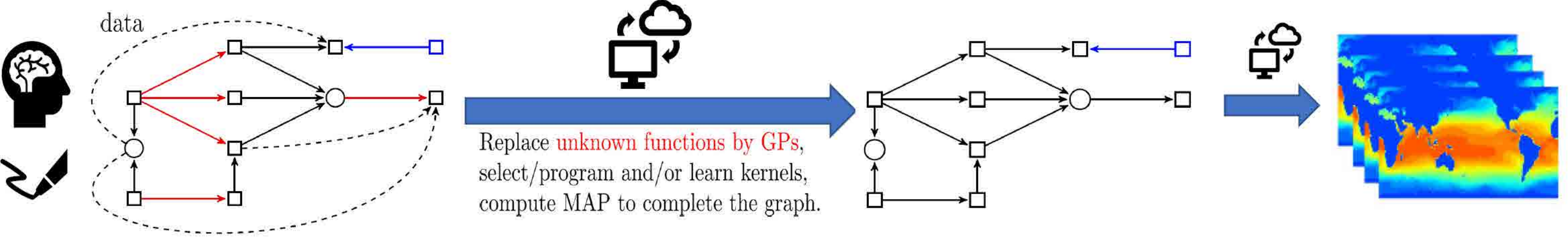}
	\end{center}

As category theory has served as a mathematical model for breaking up mathematical models into their fundamental pieces analyzed through their (functorial) relationships,    the proposed framework can serve as a model for breaking up CSE problems into their fundamental pieces and assembling them up through CGC. In that framework, creating a surrogate model for predicting the drag coefficient of a wing or extrapolating a time series could be seen as instances of the same problem (sharing the same CGC graph but solved with different kernels, variables, and data).
{ Although graphical programming environments (Simulink and Simscape are two popular ones) have facilitated the
 modeling, simulation and analysis of dynamical systems and/or physical systems, they can only solve a limited range of CSE problems because they
 require all the functions of the graph to be known.
 On the other hand, by allowing some functions of the graph to be unknown, CGC can be used as a framework for solving most CSE problems (including model/system identification) through a combination of  graph-based reasoning and   data-assimilation.}
Therefore, as calculators free us up to perform tasks of greater importance, the automation of the proposed framework has the potential to free us to perform  CSE tasks of significantly higher importance and complexity.

\subsection{Outline of the article and properties of the framework}
Sec.~\ref{seccomgra} describes the type of computational graphs used in this article. Sec.~\ref{secsamgr} introduces the notion of samples from the graph (variables of the graph satisfying its functional dependencies). Sec.~\ref{secgencgcpb} formulates the Computational Graph Completion (CGC) problem, which combines a regression problem  (approximating unknown functions) with a matrix completion problem (recovering unobserved variables in the data). Sec.~\ref{secelc} illustrates the proposed problem and framework through the identification of the Digital Twin of a nonlinear electric circuit from scarce measurements.
Sec.~\ref{secgensol} presents the proposed Gaussian Process (GP) framework for solving the general CGC problem.
This framework has several desirable properties: (1) it is highly expressive and has, therefore, a vast application scope, (2) it is interpretable and amenable to analysis, (3) its complexity reduces to that of kernel methods, (4) it is robust to data scarcity, (5) its regularization/stabilization is a rigorous and a natural generalization of the regularization strategy employed in kriging, (6) it generalizes the notion of solving linear systems of equations to that of
approximating unknown variables and functions with noisy, incomplete and nonlinear dependencies, (7) it can be automatized and can act as a pathway to identifying simple solutions to classical CSE problems.
Sec.~\ref{seckjg67g00} illustrates the flexibility and scope of the proposed framework through examples related to solving nonlinear PDEs (Sec.~\ref{seckjg67g}), learning PDEs (Sec.~\ref{seckjg67gb}), Deep Learning (Sec.~\ref{subsecekjhdedgh},  since training ANNs can naturally be formulated as a particular case of CGC, the proposed framework can naturally be employed to develop a theory for Deep Learning), the regularization of ANNs/ResNets (Sec.~\ref{secrefkwhwd88d}), dimension reduction (Sec.~\ref{subseckhey}, including nonlinear active subspace learning in Sec.~\ref{subseckhey03}), mode decomposition (Sec.~\ref{seclkjdekjdhdkjhkj}), empirical mode decomposition (Sec.~\ref{Subsectd}), empirical mode decomposition with possibly unknown non-trigonometric waveforms (Sec.~\ref{seccyc}). Sec.~\ref{Subsectd} and Sec.~\ref{seccyc} also show how the design of the graph can be made dynamic and include cycles.

\section{Computational graphs}\label{seccomgra}
We will first describe the type of graphs used here and distinguish them from knowledge graphs, belief networks, and the type of computational graphs used for backpropagation.
Knowledge graphs \cite{fensel2020knowledge} (which have become critical to many enterprises today \cite{noy2019industry}) are composed of entities (e.g., universities, countries) as nodes and relationships (e.g. \emph{is located in}) as directed edges.
Their applications include data integration, its organization, and reasoning (through knowledge graph completion \cite{lin2015learning} which consists in completing the structure of knowledge graphs by predicting missing entities or relationships).
Graphical models \cite{jordan1998learning} (also known as belief/Bayesian networks) represent
represent the conditional dependence structure between a set of random variables and enable reasoning and learning with complexity and uncertainty (through their encoded joint distribution).
Traditional computation graphs \cite{tinhofer2012computational} represent the flow of computation by organizing operations and variables into nodes linked by directed edges. They have recently become popular due to applications to backpropagation and automatic differentiation.
For our purpose, we define a computational graph as a graph representing (functional)  dependencies between a finite number of (not necessarily random) variables and functions. We will use nodes to represent variables and arrows to represent functions (in solid line) and data (in dashed line).
We distinguish nodes used to aggregate variables by drawing them as circles.
We will now introduce computational graphs through illustrative examples and then formally define such graphs.

\subsection{Functional dependencies between variables}
If a  node representing a variable $x_j$ has a single incoming solid arrow (representing a function $f$) from a  node representing a variable $x_i$, then the corresponding diagram illustrated below\\
\centerline{
\begin{tikzpicture}[->,>=stealth',shorten >=1pt,auto,node distance=3cm,
                    thick,main node/.style={rectangle,draw,font=\sffamily\Large\bfseries}]

\node[main node] (1) {$x_i$};
\node[main node] (2) [right of=1] {$x_j$};

\path[every node/.style={font=\sffamily\Large\bfseries}]
    (1) edge node [above ] {$f$} (2);

\end{tikzpicture}}
 is interpreted as the  following identity holding true
\begin{equation}
x_j=f(x_i)\,.
\end{equation}

\subsection{Sums of variables}
More generally, if a square node $j$ representing a variable $x_j$ has multiple incoming solid arrows from nodes $i$ representing other variables $x_j$,
then the identity
\begin{equation}\label{eqlkjejkdhlejk}
x_j=\sum_{i\leadsto j}  f_{i,j}(x_i)\,,
\end{equation}
holds true for all values of the $x_i$, where we write $i \leadsto  j$ for the set of nodes $i$  with an outgoing solid arrow towards $j$ and  $f_{i,j}$ for the function represented by an arrow from $i$ to $j$.
 Therefore, the following diagram\\
\centerline{
\begin{tikzpicture}[->,>=stealth',shorten >=1pt,auto,node distance=3cm,
                    thick,main node/.style={rectangle,draw,font=\sffamily\Large\bfseries}]

\node[main node] (1) {$x_i$};
\node[main node] (2) [right of=1] {$x_j$};
\node[main node] (3) [right of=2] {$x_k$};
\node[main node] (4) [below of=3, ,node distance=1.5cm] {$x_l$};

\path[every node/.style={font=\sffamily\Large\bfseries}]
    (1) edge node [above ] {$f$} (2)
    (3) edge node [above ] {$g$} (2);

\path[every node/.style={font=\sffamily\Large}]
    (4) edge node[below ] {$h$} (2);
\end{tikzpicture}}
 is  interpreted as the  identity
\begin{equation}
x_j=f(x_i)+g(x_k)+h(x_l)\,.
\end{equation}

\subsection{Aggregate variables}
We will use circles for (and only for)  variables obtained by aggregating other variables. For instance the following diagram\\
\centerline{
\begin{tikzpicture}[->,>=stealth',shorten >=1pt,auto,node distance=3cm,
                    thick,main node/.style={rectangle,draw,font=\sffamily\Large\bfseries}]

\node[main node] (1) {$x$};
\node[main node,circle] (2) [right of=1] {$z$};
\node[main node] (3) [right of=2] {$y$};

\path[every node/.style={font=\sffamily\Large\bfseries}]
    (1) edge node [above ] {$z_1$} (2)
    (3) edge node [above ] {$z_2$} (2);
\end{tikzpicture}}
 is  interpreted as the  identity
\begin{equation}
z=(x,y)\,.
\end{equation}

\subsection{Functional dependencies}\label{subeqieduhe}
Therefore, we define a computational graph as a graph representing dependencies between a finite number of variables and functions.
For instance the following diagram,\\
\centerline{
\begin{tikzpicture}[->,>=stealth',shorten >=1pt,auto,node distance=3cm,
                    thick,main node/.style={rectangle,draw,font=\sffamily\Large\bfseries}]

\node[main node] (1) {$x_1$};
\node[main node] (2)  [right of=1] {$x_2$};
\node[main node,circle] (3) [right of=2] {$x_3$};
\node[main node] (4) [right of=3] {$x_4$};
\node[main node] (5) [above of=2,node distance=1.5cm] {$x_5$};
\node[main node] (6) [above of=3,node distance=1.5cm] {$x_6$};

\path[every node/.style={font=\sffamily\Large\bfseries}]
    (1) edge node [above ] {$f_{1,2}$} (2)
    (5) edge node [above ] {$f_{5,1}$} (1)
    (2) edge node [right ] {$f_{2,5}$} (5)
    (3) edge node [right ] {$f_{3,6}$} (6)
    (2) edge node [above ] {$z_1$} (3)
    (4) edge node [above ] {$z_2$} (3);
\end{tikzpicture}}
is interpreted as the following equations holding true: $x_2=f_{1,2}(x_2)$, $x_3=(x_2,x_4)$, $x_5=f_{2,5}(x_2)$, $x_1=f_{5,1}(x_5)$, $x_6=f_{3,6}(x_3)$.
Note that, contrary to traditional computational graphs, we allow for cycles, and the cycle in the above diagram can be interpreted as
$f_{5,1}\circ f_{2,5} \circ f_{1,2}$ being the identity map.

\begin{comment}
\centerline{
\begin{tikzpicture}[->,>=stealth',shorten >=1pt,auto,node distance=3cm,
                    thick,main node/.style={rectangle,draw,font=\sffamily\Large\bfseries}]

\node[main node] (1) {$x_1$};
\node[main node] (2)  [right of=1] {$x_2$};
\node[main node,circle] (3) [right of=2] {$x_3$};
\node[main node] (4) [right of=3] {$x_4$};
\node[main node] (5) [above of=2,node distance=1.5cm] {$x_5$};
\node[main node] (6) [above of=3,node distance=1.5cm] {$x_6$};

\path[every node/.style={font=\sffamily\Large\bfseries},red]
    (1) edge node [above,red ] {$f_{1,2}$} (2)
    (5) edge node [above,red ] {$f_{5,1}$} (1);

\path[every node/.style={font=\sffamily\Large\bfseries}]
    (2) edge node [right ] {$f_{2,5}$} (5);

\path[every node/.style={font=\sffamily\Large\bfseries},red]
    (3) edge node [right,red ] {$f_{3,6}$} (6);

\path[every node/.style={font=\sffamily\Large\bfseries}]
    (2) edge node [above ] {$z_1$} (3)
    (4) edge node [above ] {$z_2$} (3);
\end{tikzpicture}}
\end{comment}

\subsection{General definition and samples from the graph}\label{secsamgr}
We generally (and formally) define a computational graph as
(1) a collection of nodes/vertices $\V$, where each vertex in $i\in \V$ represents a variable $x_i$ with values in a space $\X_i$,
(2) combined with a collections of directed edges (arrows) $\cE\subset \V \times \V$.
$\V$ is the disjoint union of $\V_{\square}$ and $\V_{\circ}$ where $\V_\square$ is the set of square-shaped vertices, and $\V_\circ$ is the set of circle-shaped vertices.
For $e\in \cE$, write $a(e)$ for the origin of the arrow $e$ and $b(e)$ for its end point.  If $b(e)$ is a square shaped vertex then $e$
represents a function $f_e$ mapping $\X_{a(e)}$ to  $\X_{b(e)}$. Furthermore if $j$ is a squared shaped vertex then the identity \eqref{eqlkjejkdhlejk} holds true.
 If $j$ is circle shaped vertex then the identity
$x_j=\otimes_{i\leadsto j} x_i$ holds true, where $\otimes_{i\leadsto j} x_i$ represents the aggregation of variables with arrow from $i$ to $j$ (in that case, $\X_j=\otimes_{i\leadsto j} \X_i$).  When necessary, the order of the aggregation is set by  ordering the arrows ending in $j$ as in Sec.~\ref{subeqieduhe}.
A collection of variables $(X_i)_{i\in \V}$ satisfying the (functional and aggregation) constraints represented by the edges of the graph is called a \emph{sample from the graph}.

\section{The Computational Graph Completion problem}\label{secgencgcpb}

Let $X_{1},\ldots,X_{N}$ be $N$ samples from the graph where each $X_s=(X_{s,i})_{i\in \V}$ is a row vector whose entries are the values taken by the variables $(x_i)_{i\in \V}$ in sample $s$. Write $X:=(X_{s,i})$ for the corresponding = $N\times \V$ matrix whose row $s$ is $X_s$.
Let $M$ be a $N\times \V$ matrix with boolean entries in $\{0,1\}$.
Write
\begin{equation}
X[M]:=\{X_{s,v}\mid M_{s,v}=1\}\,,
\end{equation}
for the subset of entries of $X$ such that $M_{s,v}=1$,
\begin{equation}
X[1-M]:=\{X_{s,v}\mid M_{s,v}=0\}\,,
\end{equation}
for the subset of entries of $X$ such that $M_{s,v}=0$.

We will now assume that some functions of the graph are unknown, write $\cE_u$ for the subset arrows representing unknown functions, and
consider the following problem, which
we will refer to as a Computational Graph Completion (CGC) problem.
\begin{Problem}\label{pbcgc}
Given $X[M]$ (a partial observation of $N$ samples from the graph)
 approximate the unknown functions $(f_e)_{e\in \cE_u}$ and the unobserved variables $X[1-M]$.
\end{Problem}
Note that Problem \ref{pbcgc} combines a regression problem (the approximation of the unknown functions $(f_e)_{e\in \cE_u}$) with a matrix completion problem (the approximation of the missing entries of $X$ in $X[M]$).
Note also that for each sample $s$,  only a  (possibly nonconstant across samples) subset of the variables may be observed/measured.
   \begin{figure}[h]
	\begin{center}
			\includegraphics[width= \textwidth]{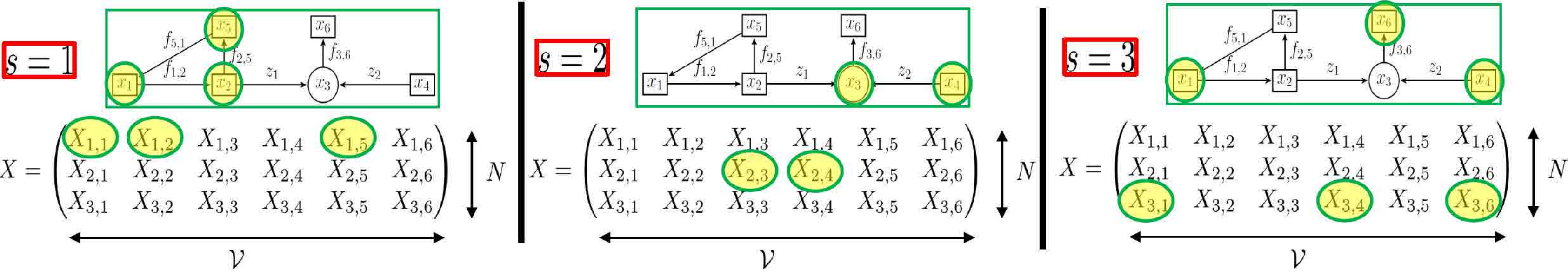}
		\caption{Observed variables in  samples $s=1,2,3$ from the graph.}\label{figM1}
	\end{center}
\end{figure}

As illustrated for the following diagram we will use red arrows to represent unknown functions and black arrows to represent known functions.\\
\centerline{
\begin{tikzpicture}[->,>=stealth',shorten >=1pt,auto,node distance=3cm,
                    thick,main node/.style={rectangle,draw,font=\sffamily\Large\bfseries}]

\node[main node] (1) {$x_1$};
\node[main node] (2)  [right of=1] {$x_2$};
\node[main node,circle] (3) [right of=2] {$x_3$};
\node[main node] (4) [right of=3] {$x_4$};
\node[main node] (5) [above of=2,node distance=1.5cm] {$x_5$};
\node[main node] (6) [above of=3,node distance=1.5cm] {$x_6$};

\path[every node/.style={font=\sffamily\Large\bfseries},red]
    (1) edge node [above,red ] {$f_{1,2}$} (2)
    (5) edge node [above,red ] {$f_{5,1}$} (1);

\path[every node/.style={font=\sffamily\Large\bfseries}]
    (2) edge node [right ] {$f_{2,5}$} (5);

\path[every node/.style={font=\sffamily\Large\bfseries},red]
    (3) edge node [right,red ] {$f_{3,6}$} (6);

\path[every node/.style={font=\sffamily\Large\bfseries}]
    (2) edge node [above ] {$z_1$} (3)
    (4) edge node [above ] {$z_2$} (3);
\end{tikzpicture}}
The following equation presents  an instance of the matrices $X$ and $M$ for $N=3$ and $|\V|=6$  in the setting of the  diagram above.
\begin{equation}
X=\begin{pmatrix}
X_{1,1}&X_{1,2}&X_{1,3}&X_{1,4}&X_{1,5}&X_{1,6}\\
X_{2,1}&X_{2,2}&X_{2,3}&X_{2,4}&X_{2,5}&X_{2,6}\\
X_{3,1}&X_{3,2}&X_{3,3}&X_{3,4}&X_{3,5}&X_{3,6}
\end{pmatrix} \quad \quad
M=\begin{pmatrix}
1&1&0&0&1&0\\
0&0&1&1&0&0\\
1&0&0&1&0&1
\end{pmatrix}
\end{equation}
Fig.~\ref{figM1} shows the corresponding observed variables for samples $s=1,2,3$.\\

\begin{figure}[h]
	\begin{center}
			\includegraphics[width= \textwidth]{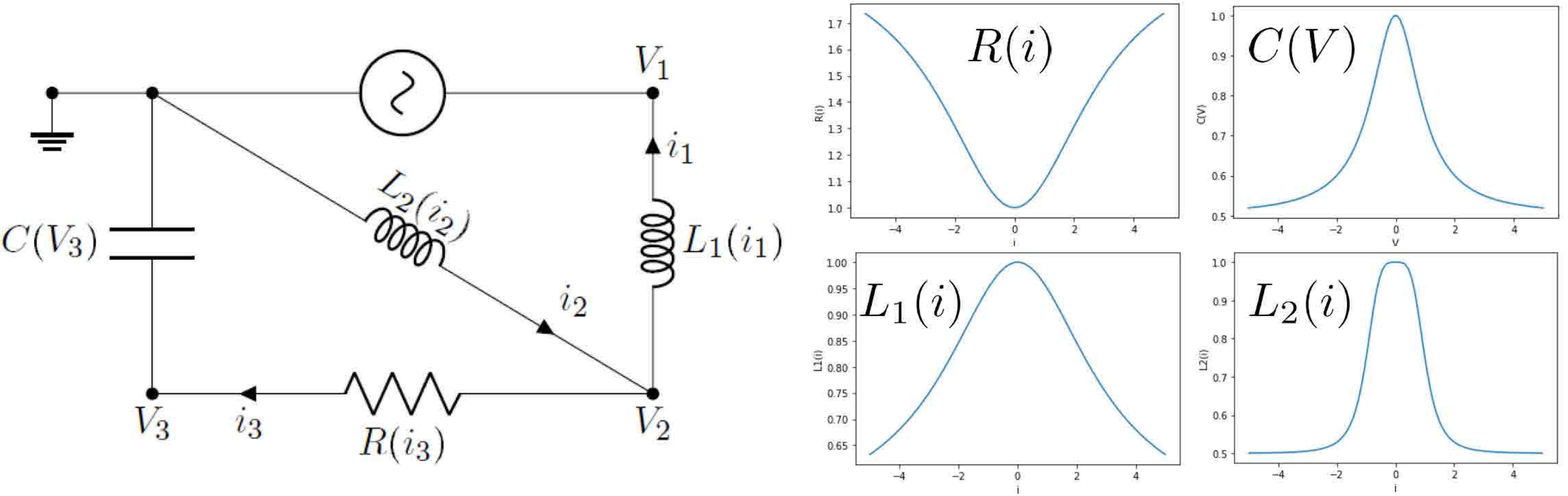}
		\caption{The nonlinear electric circuit.}\label{figcircuit1a}
	\end{center}
\end{figure}

\section{Digital twin of an electric circuit from scarce measurements}\label{secelc}
We will now motivate Problem \ref{pbcgc} by considering the problem of recovering the variables and physical laws of a nonlinear electric circuit from scarce noisy measurements (which could be seen as a simple example of Digital Twin modeling).
Our solution for this particular example will also serve as a friendly introduction to the general solution to Problem \ref{pbcgc} presented in Sec.~\ref{secgensol}.

   \begin{figure}[h]
	\begin{center}
			\includegraphics[width= \textwidth]{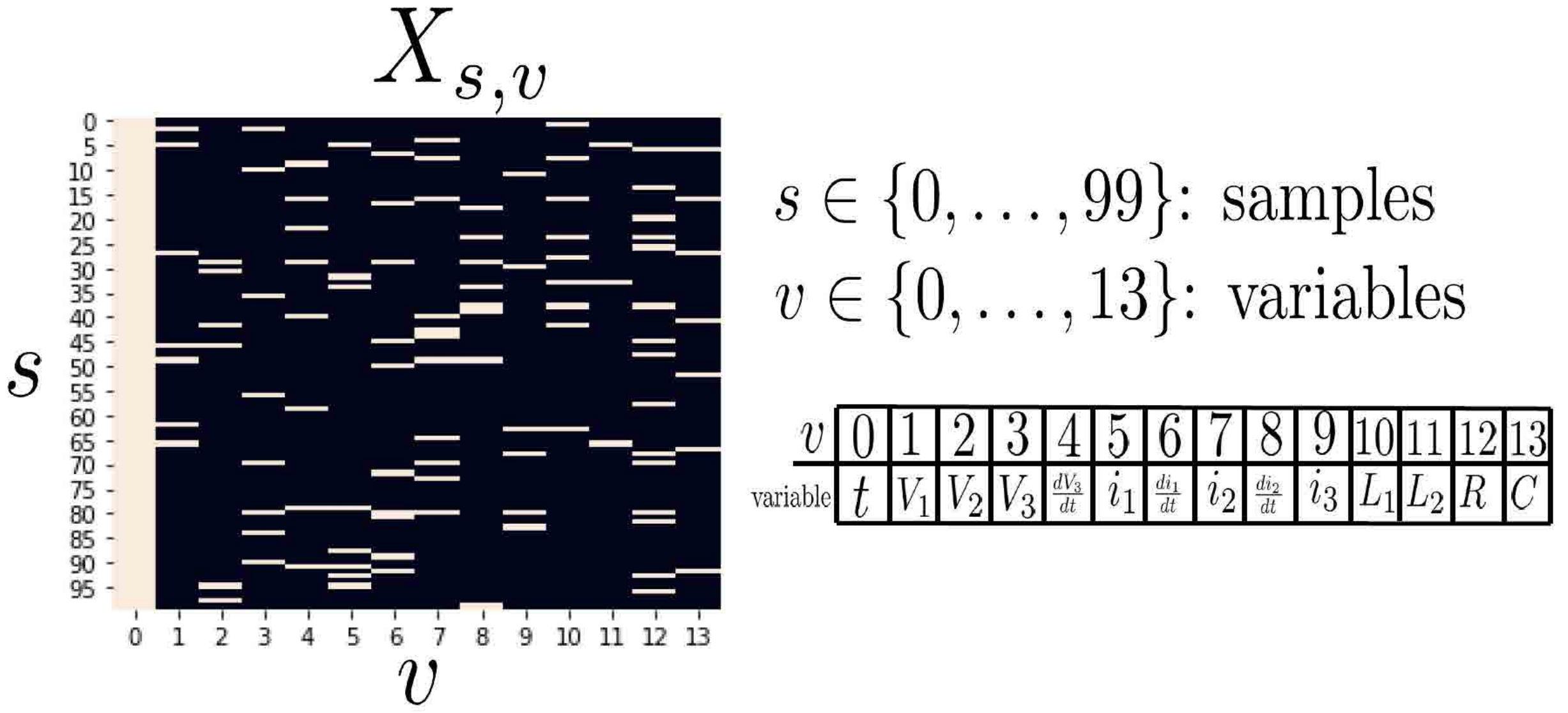}
		\caption{Observed entries of the matrix $X_{s,v}$ are colored in white.}\label{figcircuit1b}
	\end{center}
\end{figure}
\begin{figure}[h]
	\begin{center}
			\includegraphics[width= \textwidth]{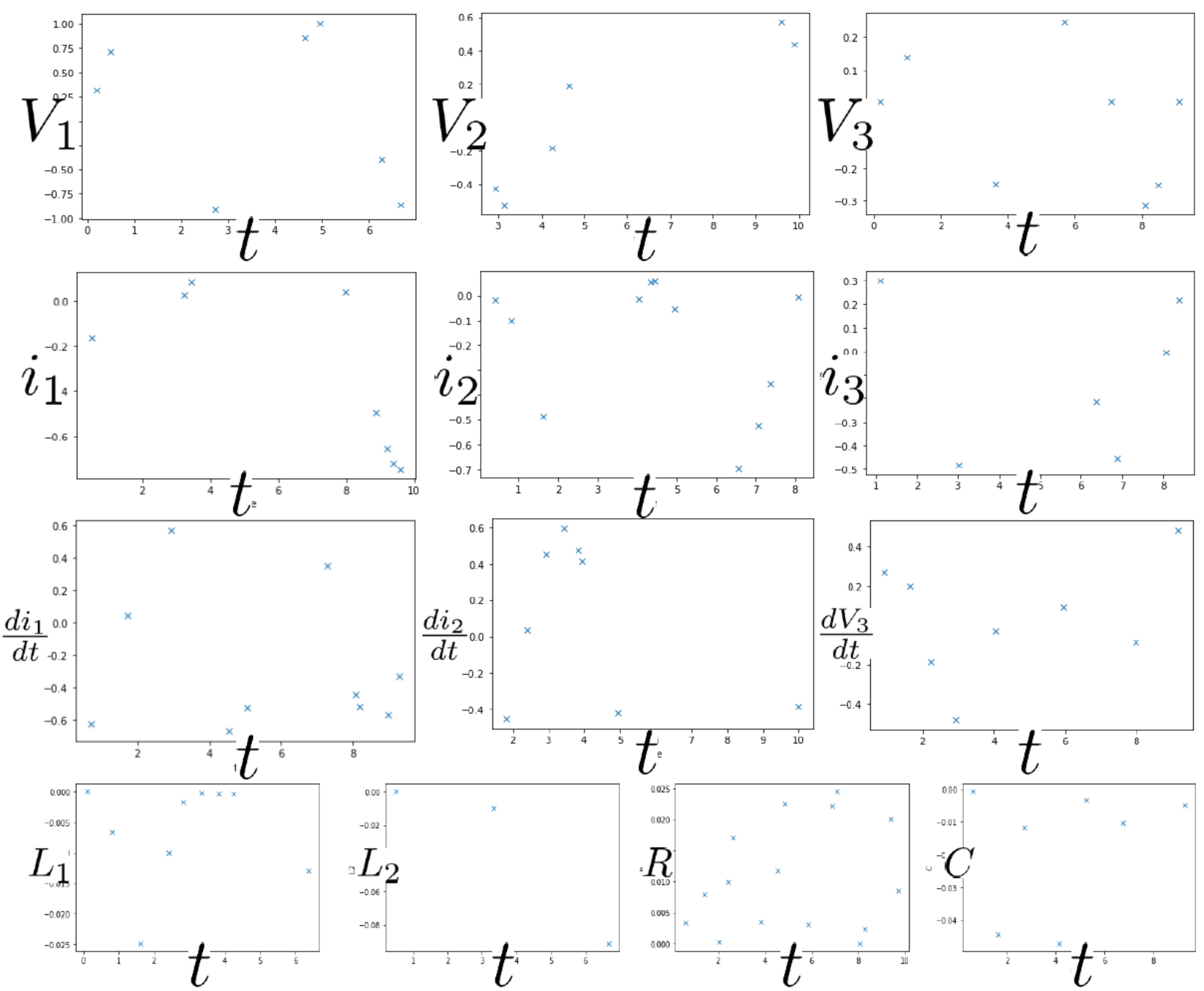}
		\caption{Values of $V_1,V_2,V_3,\frac{dV_3}{dt}, i_1,\frac{di}{dt},i_2,\frac{di_2}{dt},i_3,L_1,L_2,R,C$ at times when those variables are measured.}\label{figcircuit1c}
	\end{center}
\end{figure}

\subsection{The problem}
We consider the  electric circuit drawn in Fig.~\ref{figcircuit1a}.
We assume that this circuit is nonlinear and  that currents and voltages across the RLC components are related as shown in \eqref{eqW2}-\eqref{eqW5}.
\begin{align}
 i_1+i_3&=i_2 \label{eqW1}\\
 i_3 &=C(V_3)\frac{dV_3}{dt}\label{eqW2}\\
V_2-V_3&=R(i_3) i_3 \label{eqW3} \\
-V_2&=L_2(i_2)\frac{di_2}{dt} \label{eqW4}\\
V_2-V_1&=L_1(i_1)\frac{di_1}{dt}\label{eqW5}\,.
\end{align}
 The functions $i\rightarrow R(i), L_1(i), L_2(i)$ and $V\rightarrow C(V)$ illustrated in Fig.~\ref{figcircuit1a} are assumed to be unknown.
The time dependencies $t\rightarrow V_1(t),V_2(t),V_3(t), i_1(t),i_2(t),i_3(t)$ of all voltages and currents of the circuit  are also assumed to be unknown. The circuit is run for a given interval of time, and we seek to recover all unknown functions from scarce noisy measurements of its variables.
To describe this we  represent the state of the system by the value of the vector
\begin{equation}
x=(t,V_1,V_2,V_3,\frac{dV_3}{dt}, i_1,\frac{di}{dt},i_2,\frac{di_2}{dt},i_3,L_1,L_2,R,C)
\end{equation}
 where (see diagram  in Fig.~\ref{figcircuit1a}), $x_0=t$ represents the current time,  $x_1=V_1$ represents the value $V_1(t)$ of the voltage $V_1$ at time $x_0=t$, $x_5=i_1$ represents the value $i_1(t)$ of the current $i_1$ at time $x_0=t$,
$x_6=\frac{di_1}{dt}$ represents the value $\frac{di_1}{dt}(t)$ of the time derivative of the current $i_1$ at time $x_0=t$,
$x_{10}=L_1$ represents the value $L_1(i_1(t))$ of the inductance $L_1$ at time $x_0=t$, $x_{13}=C$ represents the value $C(V_3(t))$ of the capacitance $C$ at time $x_0=t$.
We run the circuit between $t=0$ and $t=10$ and, at times   $t_s:=\frac{s}{10}$  for $s\in \{0,1,\ldots,99\}$, observe a random subset of the values of the variables representing the state of the system.
To describe this, for $s\in \{0,\ldots,99\}$, let $(X_{s,v})_{0\leq v \leq 13}$ be the value of the vector $x$ at times $t_s$ ($X_{s,0}=t_s$). Note
that $(X_{s,v})_{0\leq s \leq 99, 0\leq v\leq 13}$  is a $100\times 14$ matrix representing the values of all the variables of the system at times $t_0,\ldots,t_{99}$.
Let $M$ be a $100\times 14$ matrix with boolean entries in $\{0,1\}$ such that $M_{s,v}=1$ if the value of $X_{s,v}$ is measured/observed and $M_{s,v}=0$ if the value of $X_{s,v}$ is not observed. For our experiments we select $M_{s,0}=1$ for all $s$ (we observe the value of the time variable at all times) and select the other entries to be independent identically distributed Bernoulli random variables such that (for $s\in \{0,\ldots,99\}$ and $v\in \{1,\ldots,13\}$)  $M_{s,v}=1$ with probability $0.07$ (e.g. for each time $t_s$ the probability that the value of $V_1$ is observed is $0.07$).

Fig.~\ref{figcircuit1b} shows the observed entries of the matrix $X_{s,v}$. Fig.~\ref{figcircuit1c} shows the values of $V_1,V_2,V_3,\frac{dV_3}{dt}, i_1,\frac{di}{dt},i_2,\frac{di_2}{dt},i_3,L_1,L_2,R,C$ at times when those variables are measured. For instance (1) the $V_1$ plot in Fig.~\ref{figcircuit1c} shows the values of the function $t\rightarrow V_1(t)$ at times $t_s$ such that $M_{s,1}=1$, and (2) the $C$ plot in Fig.~\ref{figcircuit1c} shows the values of the function $t\rightarrow C(i_3(t))$ at times $t_s$ such that $M_{s,13}=1$.
Write
\begin{equation}\label{eqljheeuidhde}
X[M]:=\{X_{s,v}\mid M_{s,v}=1\}
\end{equation}
 for the observed subset of the entries the matrix $X_{s,v}$ and consider the following problem.
\begin{Problem}\label{pbdkehgdeigd3}
Given the partial measurements $X[M]$ approximate all the entries of the matrix $(X_{s,v})$ and all the functions $t\rightarrow V_1(t),V_2(t),V_3(t),\frac{dV_3}{dt}(t), i_1(t),\frac{di}{dt}(t),i_2(t),\frac{di_2}{dt}(t),i_3(t)$, $i\rightarrow L_1(i),L_2(i),R(i)$ and
$V\rightarrow C(V)$.
\end{Problem}
Note that Problem \ref{pbdkehgdeigd3} combines a matrix completion problem with a regression problem. Furthermore, the data available for approximating each individual function is much scarcer than in traditional regression problems. For instance, as shown in Fig.~\ref{figcircuit1c}, the value of $t\rightarrow V_1(t)$ is only observed at seven points, and the proposed task is to recover $V_1$ everywhere.
Fig.~\ref{figcircuit1c} should be compared with Fig.~\ref{figcircuit2}, which illustrates the true and predicted values of all variables with the method described below.

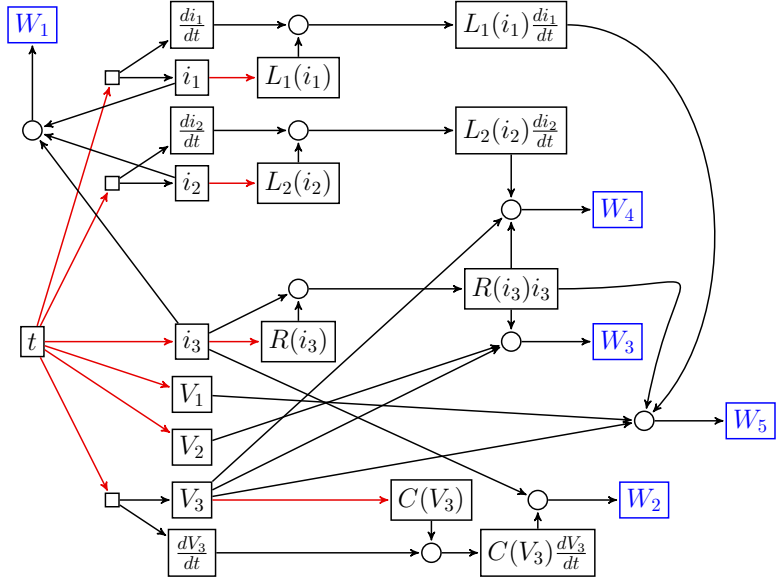
\begin{figure}
\centerline{\scalebox{0.7}{
\begin{tikzpicture}[->,>=stealth',shorten >=1pt,auto,node distance=3cm,
                    thick,main node/.style={rectangle,draw,font=\sffamily\Large\bfseries}]

\node[main node] (1) {$t$};
\node[main node] (2)  [right of=1,node distance=3cm] {$i_3$};
\node[main node] (3) [above of=2,node distance=3cm] {$i_2$};
\node[main node] (4) [above of=3,node distance=1cm] {$\frac{di_2}{dt}$};
\node[main node] (5) [above of=4,node distance=1cm] {$i_1$};
\node[main node] (6) [above of=5,node distance=1cm] {$\frac{di_1}{dt}$};
\node[main node] (7) [below of=2,node distance=1cm] {$V_1$};
\node[main node] (8) [below of=7,node distance=1cm] {$V_2$};
\node[main node] (9) [below of=8,node distance=1cm] {$V_3$};
\node[main node] (10) [below of=9,node distance=1cm] {$\frac{dV_3}{dt}$};
\node[main node] (11) [left of=3,node distance=1.5cm] {};
\node[main node] (12) [left of=5,node distance=1.5cm] {};
\node[main node] (13) [left of=9,node distance=1.5cm] {};
\node[main node] (14)  [right of=2,node distance=2cm] {$R(i_3)$};
\node[main node] (15)  [above of=14,circle,node distance=1cm] {};
\node[main node] (16)  [right of=15,node distance=4cm] {$R(i_3) i_3$};

\node[main node] (17)  [right of=9,node distance=4.5cm] {$C(V_3)$};
\node[main node] (18)  [below of=17,circle,node distance=1cm] {};
\node[main node] (19)  [right of=18,node distance=2cm] {$C(V_3) \frac{dV_3}{dt}$};

\node[main node] (20)  [right of=5,node distance=2cm] {$L_1(i_1)$};
\node[main node] (21) [above of=20,circle,node distance=1cm] {};
\node[main node] (22)  [right of=21,node distance=4cm] {$L_1(i_1) \frac{di_1}{dt}$};

\node[main node] (23)  [right of=3,node distance=2cm] {$L_2(i_2)$};
\node[main node] (24) [above of=23,circle,node distance=1cm] {};
\node[main node] (25)  [right of=24,node distance=4cm] {$L_2(i_2) \frac{di_2}{dt}$};

\node[main node] (26) [left of=4,circle,node distance=3cm] {};
\node[main node] (27) [above of=26,node distance=2cm,blue] {$W_1$};

\node[main node] (28) [above of=19,circle,node distance=1cm] {};
\node[main node] (29) [right of=28,node distance=2cm,blue] {$W_2$};

\node[main node] (30) [below of=16,circle,node distance=1cm] {};
\node[main node] (31) [right of=30,node distance=2cm,blue] {$W_3$};

\node[main node] (32) [above of=16,circle,node distance=1.5cm] {};
\node[main node] (33) [right of=32,node distance=2cm,blue] {$W_4$};

\node[main node] (34) [above of=29,circle,node distance=1.5cm] {};
\node[main node] (35) [right of=34,node distance=2cm,blue] {$W_5$};

\path[every node/.style={font=\sffamily\Large\bfseries},red]
    (1) edge node [right,red] {} (2);
\path[every node/.style={font=\sffamily\Large\bfseries},red]
    (1) edge node [right,red] {} (7);
\path[every node/.style={font=\sffamily\Large\bfseries},red]
    (1) edge node [right,red] {} (8);
\path[every node/.style={font=\sffamily\Large\bfseries},red]
    (1) edge node [right,red] {} (11);
\path[every node/.style={font=\sffamily\Large\bfseries},red]
    (1) edge node [right,red] {} (12);
\path[every node/.style={font=\sffamily\Large\bfseries},red]
    (1) edge node [right,red] {} (13);

\path[every node/.style={font=\sffamily\Large\bfseries}]
    (11) edge node [right] {} (3);
\path[every node/.style={font=\sffamily\Large\bfseries}]
    (11) edge node [right] {} (4);

\path[every node/.style={font=\sffamily\Large\bfseries}]
    (12) edge node [right] {} (5);
\path[every node/.style={font=\sffamily\Large\bfseries}]
    (12) edge node [right] {} (6);

\path[every node/.style={font=\sffamily\Large\bfseries}]
    (13) edge node [right] {} (9);
\path[every node/.style={font=\sffamily\Large\bfseries}]
    (13) edge node [right] {} (10);

\path[every node/.style={font=\sffamily\Large\bfseries},red]
    (2) edge node [right,red] {} (14);

\path[every node/.style={font=\sffamily\Large\bfseries}]
    (15) edge node [right] {} (16);
\path[every node/.style={font=\sffamily\Large\bfseries}]
    (2) edge node [right] {} (15);
\path[every node/.style={font=\sffamily\Large\bfseries}]
    (14) edge node [right] {} (15);

\path[every node/.style={font=\sffamily\Large\bfseries},red]
    (9) edge node [right,red] {} (17);

\path[every node/.style={font=\sffamily\Large\bfseries}]
    (18) edge node [right] {} (19);
\path[every node/.style={font=\sffamily\Large\bfseries}]
    (10) edge node [right] {} (18);
\path[every node/.style={font=\sffamily\Large\bfseries}]
    (17) edge node [right] {} (18);

\path[every node/.style={font=\sffamily\Large\bfseries},red]
    (5) edge node [right,red] {} (20);

\path[every node/.style={font=\sffamily\Large\bfseries}]
    (21) edge node [right] {} (22);
\path[every node/.style={font=\sffamily\Large\bfseries}]
    (6) edge node [right] {} (21);
\path[every node/.style={font=\sffamily\Large\bfseries}]
    (20) edge node [right] {} (21);

\path[every node/.style={font=\sffamily\Large\bfseries},red]
    (3) edge node [right,red] {} (23);

\path[every node/.style={font=\sffamily\Large\bfseries}]
    (24) edge node [right] {} (25);
\path[every node/.style={font=\sffamily\Large\bfseries}]
    (4) edge node [right] {} (24);
\path[every node/.style={font=\sffamily\Large\bfseries}]
    (23) edge node [right] {} (24);

\path[every node/.style={font=\sffamily\Large\bfseries}]
    (2) edge node [right] {} (26);
\path[every node/.style={font=\sffamily\Large\bfseries}]
    (3) edge node [right] {} (26);
\path[every node/.style={font=\sffamily\Large\bfseries}]
    (5) edge node [right] {} (26);
\path[every node/.style={font=\sffamily\Large\bfseries}]
    (26) edge node [right] {} (27);

\path[every node/.style={font=\sffamily\Large\bfseries}]
    (28) edge node [right] {} (29);
\path[every node/.style={font=\sffamily\Large\bfseries}]
    (2) edge node [right] {} (28);
\path[every node/.style={font=\sffamily\Large\bfseries}]
    (19) edge node [right] {} (28);

\path[every node/.style={font=\sffamily\Large\bfseries}]
    (30) edge node [right] {} (31);
\path[every node/.style={font=\sffamily\Large\bfseries}]
    (16) edge node [right] {} (30);
\path[every node/.style={font=\sffamily\Large\bfseries}]
    (8) edge node [right] {} (30);
\path[every node/.style={font=\sffamily\Large\bfseries}]
    (9) edge node [right] {} (30);

\path[every node/.style={font=\sffamily\Large\bfseries}]
    (32) edge node [right] {} (33);
\path[every node/.style={font=\sffamily\Large\bfseries}]
    (16) edge node [right] {} (32);
\path[every node/.style={font=\sffamily\Large\bfseries}]
    (25) edge node [right] {} (32);
\path[every node/.style={font=\sffamily\Large\bfseries}]
    (9) edge node [right] {} (32);

\path[every node/.style={font=\sffamily\Large\bfseries}]
    (34) edge node [right] {} (35);
\path[every node/.style={font=\sffamily\Large\bfseries}]
    (9) edge node [right] {} (34);
\path[every node/.style={font=\sffamily\Large\bfseries}]
    (7) edge node [right] {} (34);

\tikzstyle{every to}=[draw]
\draw (16) to[out=0,in=80,looseness=3,style={font=\sffamily\Large\bfseries}] node[above] {} (34);

\tikzstyle{every to}=[draw]
\draw (22) to[out=0,in=45,looseness=1,style={font=\sffamily\Large\bfseries}] node[above] {} (34);

\end{tikzpicture}}}
		\caption{The computational graph.}\label{figcircuitcgc}
\end{figure}

\subsection{Graph representation of functional dependencies between variables}\label{secgraphelec}
\begin{figure}[h]
	\begin{center}
			\includegraphics[width= \textwidth]{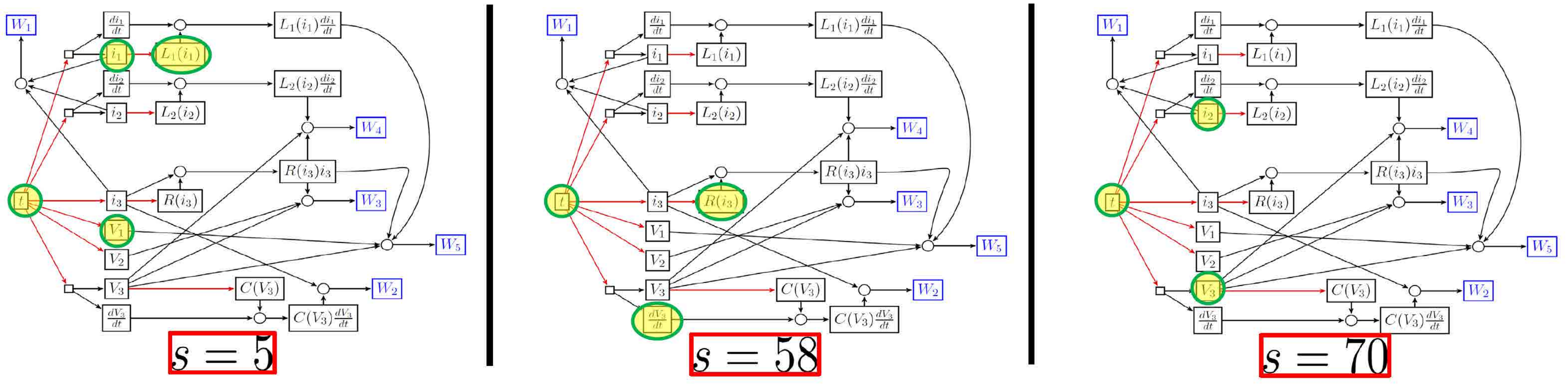}
		\caption{Observed variables in  samples $s=5,58,70$ from the graph.}\label{figM2}
	\end{center}
\end{figure}

We will now use the (imperfectly known) functional dependencies between these variables to recover them at all times. In that sense, the proposed approach could be seen as a generalization of the problem of solving a system of linear equations to that of solving a system of imperfectly known and incomplete nonlinear equations.
This is done in the proposed CGC perspective by completing the computational graph drawn in Fig.~\ref{figcircuitcgc} representing functional dependencies between variables with unknown functions (to be approximated) drawn in red.
For the circuit drawn in Fig.~\ref{figcircuit1a} these
 functional dependencies correspond to the equations \eqref{eqW1}-\eqref{eqW5}.  The variables $W_i$ illustrated in the above diagram are random variables described in Sec.~\ref{secrelwi} enabling a relaxation of the constraints imposed by equations \eqref{eqW1}-\eqref{eqW5}.
 Note that for each known value of $t$, there are  $13$ unknown variables to be approximated and only $5$ equations. Therefore the nonlinear system of equations under consideration is severely underdetermined.
 Fig.~\ref{figM2}, highlights for samples  $s=5,58,70$, the nodes of the graph drawn in Fig.~\ref{figcircuitcgc} that are associated with observed variables as indicated in  Fig.~\ref{figcircuit1b}.

\subsection{Positivity of the unknown RLC parameters}

To preserve the positivity of the unknown RLC parameters of the system, we consider the (also unknown) functions $i\rightarrow l_1, l_2, r$, and $V\rightarrow c$ defined as the log the  RLC functions, i.e.
\begin{align}
C(V_3)&=\exp(c(V_3))\label{eqW2l}\\
R(i_3) &=\exp(r(i_3)) \label{eqW3l} \\
L_2(i_2)&=\exp(l_2(i_2))\label{eqW4l}\\
L_1(i_1)&=\exp(l_1(i_1))\label{eqW5l}\,.
\end{align}
We therefore introduce the modified  matrix $(\bar{X}_{s,v})_{0\leq s \leq 99, 0\leq v\leq 13}$
defied by $\bar{X}_{s,v}:=X_{s,v}$ for $v\not \in \{10,\ldots,13\}$ and $\bar{X}_{s,v}:=\log X_{s,v}$ for $v \in \{10,\ldots,13\}$.

   \begin{figure}[h]
	\begin{center}
			\includegraphics[width= \textwidth]{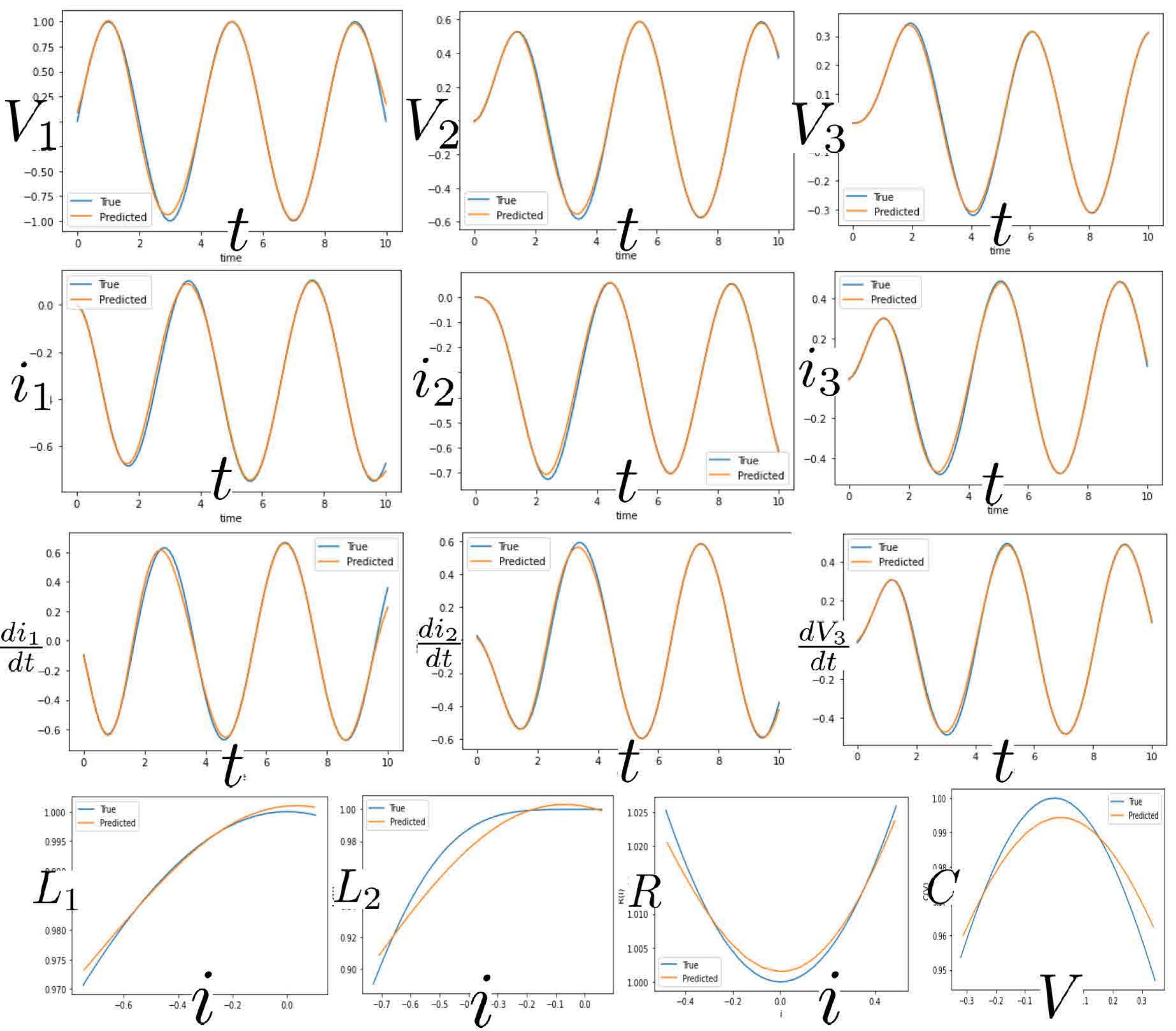}
		\caption{True values and recovered values of all variables.}\label{figcircuit2}
	\end{center}
\end{figure}

\subsection{The kernel based solution}\label{secrelwi}
Write $f:=(V_1,V_2,V_3, i_1,i_2,i_3,l_1,l_2,r,c)$ for the unknown functions to be approximated  and $Z:=(Z_{s,v})_{0\leq s \leq 99, 0\leq v\leq 13}$ for our candidate for the approximation of $\bar{X}$.  The proposed approach is then to compute $f$ and $Z$ by minimizing the  loss
\begin{equation}\label{eqkjehkdheh}
\L(f,Z):=\|f\|^2+\lambda_1 \L_1(f,Z)+\lambda_2 \L_2(Z)+\lambda_3 \big|Z[M]-\bar{X}[M]\big|^2\,,
\end{equation}
over $f$ and $Z$. In the expression \eqref{eqkjehkdheh}, $\lambda_1,\lambda_2,\lambda_3$ are strictly positive parameters chosen to be $1000$ in our numerical illustrations.
\[
\|f\|^2:=\|V_1\|_{K}^2+\|V_2\|_K^2+\|V_3\|_K^2+\| i_1\|_K^2+\|i_2\|_K^2+\|i_3\|_K^2+\|l_1\|_K^2+\|l_2\|_K^2+\|r\|_K^2+\|c\|_K^2
\]
is the sum of RKHS norms of the functions $t\rightarrow V_1,V_2,V_3, i_1,i_2,i_3$, $i \rightarrow l_1,l_2,r$ and $V\rightarrow c$ with respect to the Gaussian kernel $K(a,a'):=\exp(-|a-a'|^2)$. Write $|a|^2$ for the Euclidean norm of $a$ when $a$ is a vector and the Frobenius norm of
$a$ when $a$ is matrix/tensor/array. $\L_1(f,Z)$ is a loss enforcing the  dependencies between the unknowns functions $f$ and variables $Z$. Those dependencies are allowed to be noisy and for the sake of  clarity we did not represent all the noise variables in the computational graph drawn in Fig.~\ref{figcircuitcgc}.
\[
\begin{split}
\L_1(f,Z):=&\big|V_1(Z_{\cdot,0})-Z_{\cdot,1}\big|^2+\big|V_2(Z_{\cdot,0})-Z_{\cdot,2}\big|^2+
\big|V_3(Z_{\cdot,0})-Z_{\cdot,3}\big|^2+\big|\frac{dV_3}{dt}(Z_{\cdot,0})-Z_{\cdot,4}\big|^2
\\&+
\big|i_1(Z_{\cdot,0})-Z_{\cdot,5}\big|^2+\big|\frac{di_1}{dt}(Z_{\cdot,0})-Z_{\cdot,6}\big|^2+
\big|i_2(Z_{\cdot,0})-Z_{\cdot,7}\big|^2+
\big|\frac{di_2}{dt}(Z_{\cdot,0})-Z_{\cdot,8}\big|^2\\&+\big|i_3(Z_{\cdot,0})-Z_{\cdot,9}\big|^2+
\big|l_1(Z_{\cdot,5})-Z_{\cdot,10}\big|^2+\big|l_2(Z_{\cdot,7})-Z_{\cdot,11}\big|^2+
\big|r(Z_{\cdot,9})-Z_{\cdot,12}\big|^2\\&+\big|c(Z_{\cdot,13})-Z_{\cdot,13}\big|^2
\end{split}
\]
$\L_2(Z)$ is a loss enforcing the dependencies between the known functions and variables $Z$.
\[
\begin{split}
\L_2(Z):=&
\big|Z_{\cdot,5}+Z_{\cdot,9}-Z_{\cdot,7}\big|^2+ \big|Z_{\cdot,9}-\exp(Z_{\cdot,13})Z_{\cdot,4}\big|^2
+ \big|Z_{\cdot,2}-Z_{\cdot,3}-\exp(Z_{\cdot,12})Z_{\cdot,9}\big|^2\\&+
\big|Z_{\cdot,2}+\exp(Z_{\cdot,11})Z_{\cdot,8}\big|^2
+\big|Z_{\cdot,2}-Z_{\cdot,1}-\exp(Z_{\cdot,10})Z_{\cdot,6}\big|^2\,.
\end{split}
\]
Note that the dependencies represented by $\L_2$ are also noisy and correspond to replacing \eqref{eqW1}-\eqref{eqW5} by the following equations in which the $W_i$ (illustrated in the computational graph drawn in Fig.~\ref{figcircuitcgc}) are independent centered Gaussian random variables with variance $1/\lambda_2$.
\begin{align}
 i_1+i_3-i_2&=W_1 \label{eqW1b}\\
 i_3 -C(V_3)\frac{dV_3}{dt}&=W_2\label{eqW2b}\\
V_2-V_3-R(i_3) i_3&=W_3 \label{eqW3b} \\
-V_2-L_2(i_2)\frac{di_2}{dt}&=W_4 \label{eqW4b}\\
V_2-V_1-L_1(i_1)\frac{di_1}{dt}&=W_5\label{eqW5b}\,.
\end{align}
Using the notations of \eqref{eqljheeuidhde},
 $\big|Z[M]-\bar{X}[M]\big|^2$ is the Euclidean norm of the difference between the $Z[M]$ and $\bar{X}[M]$. Note the last term in \eqref{eqkjehkdheh} enforces the constraints imposed by the observation of the data under the assumption that the data has been corrupted by i.i.d. noise (of centered Gaussian distribution with variance $1/\lambda_3$).

 \subsection{Reduction to a finite-dimensional optimization problem and numerical solution}
 For a given $Z$, \eqref{eqkjehkdheh} is quadratic in $f$ and its minimizer (in $f$) can be computed explicitly as  described in Sec.~\ref{seckejddjdhkj} (as a function of $Z$), e.g. (using vectorized notations) writing $\phi^1=(\updelta_{Z_{\cdot,0}})$, $\phi^2=(\updelta_{Z_{\cdot,0}}\circ \frac{d}{dt})$, $\phi=(\phi^1,\phi^2)$, the minimizer in $i_1$ is $i_1=K(\cdot,\phi) (K(\phi,\phi)+\lambda_1^{-1} I)^{-1} (Z_{\cdot,5},Z_{\cdot,6})$ and the value of $\|i_1\|_K^2+\lambda_1\big(\big|i_1(Z_{\cdot,0})-Z_{\cdot,5}\big|^2+\big|\frac{di_1}{dt}(Z_{\cdot,0})-Z_{\cdot,6}\big|^2\big)$ at the minimum in $i_1$ is $(Z_{\cdot,5},Z_{\cdot,6})^T(K(\phi,\phi)+\lambda_1^{-1} I)^{-1} (Z_{\cdot,5},Z_{\cdot,6})$.
 The proposed strategy is then to minimize \eqref{eqkjehkdheh}  over $f$ first and then over $Z$ next.
 If discrete times $Z_{\cdot,0}$ form a fine enough mesh of the time interval $[0,10]$ then  minimizers of $Z\rightarrow  \min_{f}\L(f,Z)$ are well approximated by substituting $Z_{\cdot,4}, Z_{\cdot,6}, Z_{\cdot,8}$ (which represent the values of the time derivatives of $V_3, i_1$ and $i_2$ at the discrete time points) with\\
  $Z_{\cdot,4}=K(\phi^2,\phi^1)K(\phi^1,\phi^1)^{-1} Z_{\cdot,3}$, $Z_{\cdot,6}=K(\phi^2,\phi^1)K(\phi^1,\phi^1)^{-1} Z_{\cdot,5}$ and\\
 $Z_{\cdot,8}=K(\phi^2,\phi^1)K(\phi^1,\phi^1)^{-1} Z_{\cdot,7}$ (which are the values of the time derivatives of $V_3, i_1$ and $i_2$ at the discrete time points obtained by simply regressing the  values of $V_3, i_1$ and $i_2$). We have used this technique here and
 Fig.~\ref{figcircuit2} shows the true and recovered values of all functions and variables of the system (in the time interval $[0,t]$ for the functions $t\rightarrow V_1,V_2,V_3, i_1,i_2,i_3$,  in the range spanned by (a) $i_1$ for $L_1$, (b)  $i_2$ for $L_2$, (c) $i_3$ for $R$, and (d) $V_3$ for $C$).

\section{General solution to Computational Graph Completion problems}\label{secgensol}
We will now describe the  Gaussian Process (GP) framework for solving
 Problem \ref{pbcgc}. This framework will allow us to generalize the particular strategy used in Sec.~\ref{secelc}.

\subsection{Computational Graph Completion with Gaussian Processes}
 Our approach to solving CGC problems can be summarized as (1) replacing unknown functions by  GPs, and (2) approximating those functions by the MAP estimator of those GPs given available data. The covariance kernels of these GPs may be pre-determined by the user (based on prior information on regularity/symmetries) or learned from data (e.g., using Kernel Flows as described in \cite{owhadi2019kernel, hamzi2021learning}).
We will focus this manuscript on the situation where those kernels are given. We assume that the nodes of our computational graphs represent variables in separable Hilbert spaces and employ GPs with scalar/vector/operator-valued kernels as covariance kernels (see \cite{owhadinotesopvkgp2021} for a short reminder on operator-valued kernels, their feature maps, and corresponding Gaussian Processes).
If $f$ is a function mapping the variable $x$ to the variable $y$,  we may  not specify the (finite or infinite dimensional) Hilbert spaces $\X$ and $\Y$ containing $x$ and $y$, nor the underlying (possibly operator valued) kernel\footnote{Write $\L(\Y)$ for  the set of bounded linear operators mapping $\Y$ to $\Y$} $K\,:\, \X\times \X\rightarrow \L(\Y)$  used as a covariance kernel for the GP randomization of $f$.
Indeed, one motivation for the proposed framework is to enable high-level computation and reasoning with GPs, and abstract away/automatize the underlying technicalities.
Thus, prior to presenting our general solutions to Problem \ref{pbcgc} we will first set notations by formulating and solving classical interpolation and regression problems as CGC problems with GPs.

\subsection{Jointly measured/observed variables}
If for some $s\in \{1,\ldots,N\}$, two and only two variables $X_{s,i}$ and $X_{s,j}$ are observed then we draw a dashed arrow between nodes $i$ and $j$.
 If $X_{s,j}=g(X_{s,i})$ where $g$ is a known function then we may label that arrow with $g$ (to represent the corresponding data) and point the arrow from $i$ to $j$. Otherwise, we may label the dashed arrow with the corresponding data.

\subsection{Interpolation}\label{secinterpset}
The simplest example of computation graph completion problem, illustrated in the following diagram, is that of identifying an unknown function $f$ from data $(X_i,Y_i=f(X_i))_{i=1,\ldots,N}$.\\
\centerline{
\begin{tikzpicture}[->,>=stealth',shorten >=1pt,auto,node distance=3cm,
                    thick,main node/.style={rectangle,draw,font=\sffamily\Large\bfseries}]

\node[main node] (1) {$x$};
\node[main node] (2) [right of=1] {$y$};

\path[every node/.style={font=\sffamily\Large\bfseries},red]
    (1) edge node [above ] {$f$} (2);

\path[every node/.style={font=\sffamily\Large}]
    (1) edge  [bend right,dashed ] node[below ] {$(X,Y)$} (2);

\end{tikzpicture}}
Given a kernel $K$, the proposed GP approach is the classical GP interpolation one of approximating $f$ with
 the conditional expectation
\begin{equation}
f^\dagger(x)=\E[\xi|\xi(X)=Y]
\end{equation}
 where $\xi\sim \N(0,K)$ is a centered GP with covariance kernel $K$, $X=(X_1,\ldots,X_N)$, $Y=(Y_1,\ldots,Y_N)$ and $\xi(X):=\big(\xi(X_1),\ldots,\xi(X_N)\big)$. Writing $\H_K$ and $\|\cdot\|_K $ for the RKHS space and norm associated with $K$, recall (see appendix) that $f^\dagger$ can also be identified as the minimizer  of
\begin{equation}\label{eqhgvygvgyv}
\begin{cases}
\minimize_{f\in \H_K}&\|f\|_{K}^2\\
\st&f(X)=Y\,,
\end{cases}
\end{equation}
and admits the representer formula
 \begin{equation}\label{eqhgvygvgyv2}
 f^\dagger(\cdot)= K(\cdot,X)  K(X,X)^{-1} Y\,,
 \end{equation}
where $K(\cdot,X)$ is the $N$-vector with entries $K(\cdot,X_i)$ and $K(X,X)$ is the $N\times N$ matrix with entries $K(X_i,X_j)$.

\subsection{Random variables}
Variables may be primal (black square),  aggregation of other variables (black circle), or random (blue square).
We write $\V_r$ for the set of nodes representing random variables.
Using the notations of Sec.~\ref{secgencgcpb}, we assume that the random variables $(X_{s,i})_{1\leq s \leq N, i\in \V_r}$ are independent centered Gaussian random variables/vectors/fields with covariances matrices/operators $(K_i)_{i\in \V_r}$. Although we require independence for ease of presentation, this assumption imposes no limitation since correlations and other distributions can be considered by pushing forward Gaussian vectors through known functions.

\subsection{Regression}
The following diagram illustrates  the computation graph completion problem of identifying an unknown function $f$ given noisy observations $(X_i,Y_i)_{i=1,\ldots,N}$ of $f$, where $Y_i=f(X_i)+W_i$ and the $W_i$ are i.i.d. $\cN(0,\sigma^2 I_\Y)$ random variables. Recall that we draw random variables in blue. We will use random variables (noise) to regularize the recovery/approximation of unknown functions from data.
The function associated with the arrow connecting $w$ to $y$ is the identity function which omit as a label to simplify the display of our diagrams.\\
\centerline{
\begin{tikzpicture}[->,>=stealth',shorten >=1pt,auto,node distance=3cm,
                    thick,main node/.style={rectangle,draw,font=\sffamily\Large\bfseries}]

\node[main node] (1) {$x$};
\node[main node] (2) [right of=1] {$y$};
\node[main node] (3) [above of=2,blue, node distance=1cm] {$w$};

\path[every node/.style={font=\sffamily\Large\bfseries},red]
    (1) edge node [above ] {$f$} (2);

\path[every node/.style={font=\sffamily\Large\bfseries}]
    (3) edge node [above ] {} (2);

\path[every node/.style={font=\sffamily\Large}]
    (1) edge  [bend right,dashed ] node[below ] {$(X,Y)$} (2);

\end{tikzpicture}}
Given a kernel $K$, the proposed approach is then to consider the GP $\xi \sim \cN(0,K)$ and approximate
 $f$ with
 the conditional expectation
\begin{equation}\label{eqlwjkdejd}
f^\dagger(x)=\E[\xi|\xi(X)+W=Y]
\end{equation}
where $W$ is the $N$-vector with entries $W_i$. \eqref{eqlwjkdejd} can be identified as the minimizer of
\begin{equation}\label{eqhgvwygvgyv}
\begin{cases}
\minimize_{f\in \H_K,W\in \Y^N}&\|f\|_{K}^2+\frac{1}{\sigma^2}\|W\|^2_{\Y^N}\\
\st & f(X)+W=Y\,,
\end{cases}
\end{equation}
i.e. (using the constraint to eliminate $W$), the minimizer  of,
\begin{equation}\label{eqhgvygvgyvs2}
\minimize_{f\in \H_K}\|f\|_{K}^2+\frac{1}{\sigma^2}\|f(X)-Y\|^2_{\Y^N}\,,
\end{equation}
which admits the following representer formula,
\begin{equation}\label{eqajkjwdhjbdjeh}
f^\dagger(x)=K(x,X)\big(K(X,X)+\sigma^2 I\big)^{-1} Y\,.
\end{equation}

\begin{Remark}\label{rmkmot1}
Although this (classical) approach may appear simple, it can be highly effective when the underlying kernel is also learned from data \cite{hamzi2021learning, hamzi2021simple}. Considering the extrapolation of time series obtained from satellite data as an example, this simple data-adapted kernel  perspective outperforms
 (both in complexity and accuracy) PDE-based and ANN-based methods \cite{hamzi2021simple}.
\end{Remark}

\subsection{General CGC solution with GPs}\label{seccgcov}
Consider the general CGC problem \ref{pbcgc}. The proposed GP solution is to randomize the unknown functions $(f_e)_{e\in \cE_u}$ with independent\footnote{Evidently, the independence assumption can be relaxed.} centered GPs with kernels $(K_e)_{e\in \cE_u}$ and approximate those unknown functions with a MAP estimator given the data.
For $j\in \V$ write $X_{\cdot,j}$ for the column $j$ of the matrix $X$ ($(X_{s,j})_{1\leq s \leq N}$).
For a function $f$ with domain $\X_j$ write $f(X_{\cdot,j})$ for the column vector with entries $(f(X_{s,j}))_{1\leq s \leq N}$.
For $e\in \cE$, write  $a(e)$ is the node at the origin of the arrow $e$. For $i\in \V_r$ and $x_i\in \X_i$, write $\|x_i\|_{K_i}^2:= x_i^T K_i^{-1} x_i $ for the (RKHS) norm defined by the covariance matrix/operator $K_i$.
Write $\V_\square^\iota$ for the set of square nodes with at least one incoming arrow. For $e=(i,j)\in \cE$ write $f_e$ or $f_{i,j}$ for the function represented by $e$. For $j\in \V$, write $e\leadsto j$ for the set of arrows $e\in \cE$ ending in $j$.
Then the proposed GP solution to the CGC problem is to approximate the unknown functions $(f_e)_{e\in \cE_u}$ and the matrix $X$ with
minimizers $(f_e^\dagger)_{e\in \cE_u}$ and  $Z^\dagger$ of the variational problem.
\begin{equation}\label{eqhdewdgvgyv}
\begin{cases}
\minimize_{(f_e)_{e\in \cE_u}, Z, (Y_e)_{e\in \cE}}&\sum_{e\in \cE_u}\|f_e\|_{K_e}^2+\sum_{s=1}^N \sum_{i\in \V_r}\|Z_{s,i}\|^2_{K_i}\\
\st & f_e(Z_{\cdot,a(e)})=Y_{\cdot,e} \,,\text{ for }e\in \cE\,,\\
{\rm and }& Z_{\cdot,j}=\sum_{e\leadsto j} Y_{\cdot,e}\,,\text{ for } j\in \V_\square^\iota\,, \\
{\rm and }& Z_{\cdot,j}=\otimes_{i\leadsto j} Z_{\cdot,i}\,,\text{ for } j\in \V_\circ\,,\\
{\rm and }& Z[M]=X[M]\,.
\end{cases}
\end{equation}
where the minimization is over $(f_e)_{e\in \cE_u}\in \prod_{e\in \cE} \H_{K_e}$, $Z=\{Z_{s,i}\in \X_i\mid 1\leq s \leq N,\, i\in \V\}$
and $(Y_{s,e})_{1\leq s \leq N, e\in \cE}$ (with $Y_{s,e}\in  \cR(f_e)$, writing $\cR(f_e)$ for range of $f_e$).
Minimizing over $(f_e)_{e\in \cE_u}$ first in \eqref{eqhdewdgvgyv} we obtain the following theorem turning the MAP estimation problem into a computationally tractable problem.
\begin{Theorem}
$(f_e)_{e\in \cE_u}$, $Z$  and the $Y_{\cdot,e}$ are a minimizer of \eqref{eqhdewdgvgyv} if and only if
\begin{equation}\label{eqeersedgvgyv}
f_e(\cdot)=K_e(\cdot, Z_{\cdot,a(e)}^\dagger) K_e(Z_{\cdot,a(e)}^\dagger,Z_{\cdot,a(e)}^\dagger)^{-1}Y_e^\dagger
\end{equation}
where $(Y_e^\dagger)_{e\in \cE}$ and $Z^\dagger$ are a minimizer of
\begin{equation}\label{eqhdeeedgyv}
\begin{cases}
\minimize_{Z, (Y_e)_{e\in \cE_u}}&\sum_{e\in \cE_u}Y_e^T K_e(Z_{\cdot,a(e)},Z_{\cdot,a(e)})^{-1}Y_e+\sum_{s=1}^N \sum_{i\in \V_r}\|Z_{s,i}\|^2_{K_i}\\
\st& f_e(Z_{\cdot,a(e)})=Y_{\cdot,e} \,,\text{ for }e\in \cE/\cE_u\,,\\
{\rm and }& Z_{{\cdot,j}}=\sum_{e\leadsto j} Y_{\cdot,e}\,,\text{ for } j\in \V_\square^\iota \,, \\
{\rm and }& Z_{{\cdot,j}}=\otimes_{i\leadsto j} Z_{\cdot,i}\,,\text{ for } j\in \V_\circ\,,\\
{\rm and }& Z[M]=X[M]\,.
\end{cases}
\end{equation}
\end{Theorem}

\subsection{Optimal recovery and gamblets}\label{seckejddjdhkj}
Consider a separable Banach space of functions $\B$  and its dual $\B^*$ with their duality pairing denoted by $[ \cdot, \cdot ]$. If $\B$ is a space of functions  $u\,:\, \X\rightarrow \Y$ then each element of $\B$ can be extended to a linear functional mapping $\phi \in \B^*$ to $[\phi,u]\in \R$.
This simple   observation leads to a generalization of CGC to computational graphs with variables in $\B^*$. To that end consider the problem, illustrated in the following diagram, of recovering an unknown element $u\in \B$ (seen as a function mapping $\varphi\in \B^*$ to $u(\varphi):=[\varphi,u]\in \R$) given the observation of $(Y_i:=[\phi_i,u])_{i=1,\ldots,N}\in \R^N$ where the $\phi_i$ are $N$ linearly independent elements of $\B^*$.

\begin{equation}\label{eqkhetftydgy}
\begin{tikzpicture}[->,>=stealth',shorten >=1pt,auto,node distance=3cm,
                    thick,main node/.style={rectangle,draw,font=\sffamily\Large\bfseries}]

\node[main node] (1) {$\varphi$};
\node[main node] (2) [right of=1] {$[\varphi,f]$};

\path[every node/.style={font=\sffamily\Large\bfseries},red]
    (1) edge node [above ] {$u$} (2);

\path[every node/.style={font=\sffamily\Large}]
    (1) edge  [bend right,dashed ] node[below ] {$(\phi,Y)$} (2);

\end{tikzpicture}
\end{equation}

To define the proposed GP solution to this CGC problem, we introduce a covariance operator \cite[Chap.~11]{owhadi2019operator} defined as a linear bijection
 $K: \B^* \to \B$ that is symmetric  ($[ \phi,K\varphi]=[\varphi,K\phi]$) and positive ($[ \phi, K \phi]>0$ for $\phi \ne 0$).
 If the elements of $\B$ are continuous functions then, for ease of notations, we also write $K(x,x'):=[\updelta_x, K \updelta_{x'}]$ for the kernel associated with $K$.
 Consider the GP $\xi\sim \cN(0,K)$ defined (\cite[Chap.~17]{owhadi2019operator} and \cite[Sec.~8]{owhadi2020ideas}) as an isometry from $\B^*$ to a Gaussian space mapping each   $\varphi\in \B^*$ to a centered scalar valued Gaussian random variable with variance $[\varphi,K\varphi]$.
 The GP solution to this CGC problem is then to approximate $f$ with the conditional expectation (writing $\xi(\phi)$ for the vector with entries $\xi(\phi_i)$)
 \begin{equation}\label{eqleddejk}
 u^\dagger=\E[\xi|\xi(\phi)=Y]\,.
 \end{equation}
Writing
$\|u\|_K^2:=[K^{-1}u,u]
$ for the quadratic norm defined by $K$ on $\B$, $u^\dagger$ can also be identified as the minimizer  of (writing $u(\phi)$ for the vector with entries $[\phi_i,u]$)
\begin{equation}\label{eqhgvygvgewew2yv}
\begin{cases}
\minimize_{u\in \B}&\|u\|_{K}^2\\
\st&u(\phi)=Y\,,
\end{cases}
\end{equation}
and admits the representer formula
 \begin{equation}\label{eqhgvygvgyv2e}
 [\varphi,u^\dagger]= K(\varphi,\phi)  K(\phi,\phi)^{-1} K\phi\,
 \end{equation}
where $K \phi$  is the $N$-vector with entries $K\phi_i$, $K(\phi,\phi)$ is the $N\times N$ matrix with entries $[\phi_i,K \phi_j]$ and
$ K(\varphi,\phi) $ is the $N$-vector with entries $[\varphi,K \phi_i]$.
\begin{Remark}
A particular case of the example considered here is $\B=\H^s_0(\Omega)$  (writing $H^s_0(\Omega)$ for the closure of the set of smooth functions with compact support in $\Omega$ with respect to the Sobolev norm $\|\cdot\|_{H^s(\Omega)}$), with its dual
  $\B^*=H^{-s}(\Omega)$ defined by the pairing   $[\phi,v]:=\int_{\Omega}{\phi u}$ obtained from the  Gelfand triple
 $H^s(\Omega) \subset L^2(\Omega) \subset H^{-s}(\Omega)$  \cite[Chap.~2]{owhadi2019operator}.
Given an elliptic operator $\mathcal{L}\,:\, H^s_0(\Omega)\rightarrow H^{-s}(\Omega)$ (i.e., a linear bijection that is positive and symmetric in the sense that $\int_\Omega u \mathcal{L} u\geq 0$ and $\int_\Omega u \mathcal{L} v= \int_\Omega v \mathcal{L} u$) let $K$ be the solution map $\L^{-1}$.
Let the $\phi_i$ be (1) (for $s>d/2$) delta Dirac functions $\updelta_{x_i}$ centered at $N$ homogeneously distributed points $x_i$ of $\Omega$, or (2) (for $s>0$) indicator functions of subsets $\tau_i$ forming a regular  partition of $\Omega$ \cite[Sec.~4.3]{owhadi2019operator}.
Then the minimizer of \eqref{eqhgvygvgewew2yv} is
 \begin{equation}\label{eqlededdejk}
 u^\dagger=\sum_{i=1}^n [\phi_i,u] \psi_i
 \end{equation}
where the elements $\psi_i\in \B$ (defined by  $\psi_i=\sum_{i=1}^N K(\phi,\phi)^{-1}_{i,j} K\phi_j$)
are optimal recovery splines \cite{micchelli1977survey} or gamblets \cite{Owhadi:2014, owhadi2017multigrid} adapted to the operator $\L$ (i.e., operator adapted wavelets \cite{owhadi2019operator} that are both localized and near optimal in approximating the solution space  $\L^{-1} L^2(\Omega)$).
We refer to \cite{owhadi2019operator, OwhScoSchNotAMS2019} for a further survey of interplays between numerical approximation and statistical inference.
\end{Remark}

\subsection{Computational Graph Completion with generalized variables}\label{secgenvar}
The setting of Sec.~\ref{seckejddjdhkj} allows us to generalize CGC to situations where some of the variables $x_i$ are vectors whose entries are linear functionals action on spaces of functions, i.e. $\X_i=(\B_e^*)^m$ where $e$ is an edge originating from $i$, $\B_e^*$ is the dual space of space of functions,  $e$ is representing an element of $f_e \in \B_e$ and, for $x_i=(x_{i,1},\ldots,x_{i,m})$, $f_e(x_i)$ is interpreted as the vector $([x_{i,1},f_e],\ldots,[x_{i,m},f_e])$. In that situation, the results  and formulas of Sec.~\ref{seccgcov} hold true with the following alterations: if $f_e$ is unknown and acting on linear functionals, then the $K_e$ describing its GP randomization is a covariance operator and for
$x_i,x_i'\in(\B_e^*)^m$, $K_e(x_i,x_i')$ is a matrix with entries $[x_{i,j}, K_e x_{i,l}']$.
In that setting an unknown function mapping $x$ to $(u(x),\Delta u (x))$ should be displayed as\\
 \centerline{
\begin{tikzpicture}[->,>=stealth',shorten >=1pt,auto,node distance=3cm,
                    thick,main node/.style={rectangle,draw,font=\sffamily\Large\bfseries}]

\node[main node] (1) {$x$};
\node (2) [main node, right of=1, node distance=4cm] {$(\updelta_x,-\updelta_x \circ \Delta) $};
\node (3) [main node, right of=2, node distance=6cm] {$(u(x),-\Delta u(x))$};
\path[every node/.style={font=\sffamily\Large\bfseries}]
    (1) edge node [above,red] {} (2);
\path[every node/.style={font=\sffamily\Large\bfseries},red]
    (2) edge node [above,red] {$u$} (3);
\end{tikzpicture}
}
but for ease of presentation, we will simplify such diagrams by drawing them as follows.\\
 \centerline{
\begin{tikzpicture}[->,>=stealth',shorten >=1pt,auto,node distance=3cm,
                    thick,main node/.style={rectangle,draw,font=\sffamily\Large\bfseries}]

\node[main node] (1) {$x$};
\node (2) [main node, right of=1, node distance=4cm] {$(u(x),-\Delta u(x))$};
\path[every node/.style={font=\sffamily\Large\bfseries},red]
    (1) edge node [above,red] {$u$} (2);
\end{tikzpicture}
}

\subsection{Relaxation and regularization with random variables}\label{secrelcgc}
A particular case of the solution presented in Sec.~\ref{seccgcov} and \ref{secgenvar} is to use random variables to regularize the underlying constraints imposed by unknown functions, known functions, and measurements.
As shown in \cite{owhadi2020ideas} (see also Sec.~\ref{secrefkwhwd88d}),  from the perspective of kriging with composed GPs/kernels, relaxing output constraints with nuggets (small Gaussian noise capturing the variability of the data \cite{cressie1988spatial}) is not sufficient to ensure stability, one must also relax all composition steps\footnote{This type of regularization was introduced in \cite{owhadi2020ideas} (as a principled and rigorous alternative to Dropout \cite{srivastava2014dropout}) for ANNs/ResNets and their kernelized variants where it was shown to be necessary and sufficient to avoid regression instabilities (the lack of stability/continuity of regressors with respect to data and input variables).}.
To describe this consider (for ease of presentation) Problem \ref{pbcgc} in the setting where there are no random variables $\V_r=\empty$.
Let $f:=(f_e)_{e\in \cE_u}$ represent all unknown functions and let $g:=(f_e)_{e\in \cE/\cE_u}$ represent all known functions.
Then, given strictly positive relaxation parameters $\lambda_{i,j}>0$ (described below), the proposed solution  to Problem \ref{pbcgc} is to
recover $f$ and $X$ with a minimizer of the following variational problem
\begin{equation}\label{eqhdewdgvgyvb}
\begin{cases}
\minimize_{f,Z, Y} &\|f\|^2+ \L_1(f, Y,Z)+\L_2( Y,Z)+\L_3(Z[M],X[M])\,,\\
\st &  Z_{{\cdot,j}}=\otimes_{i\leadsto j} Z_{\cdot,i}\,,\text{ for } j\in \V_\circ\,.
\end{cases}
\end{equation}
where $\|f\|^2:=\sum_{e\in \cE_u}\|f_e\|_{K_e}^2$ is the sum of the RKHS norms of all unknown functions,
\begin{equation}
\L_1(f,Z):=\sum_{e\in \cE} \lambda_{1,b(e)}\|f_e(Z_{\cdot,a(e)})-Y_{e}\|_{\X_{b(e)}^N}^2\,,
\end{equation}
relaxes dependencies between known and unknown functions, $Y$ and $Z$,
\begin{equation}
\L_2(Z,Y):=  \sum_{j\in \V_\square^\iota} \lambda_{2,j} \|Z_{{\cdot,j}}-\sum_{e\leadsto j} Y_e\|_{\X_{j}^N}^2 \,, \\
\end{equation}
relaxes dependencies between $Y$ and $Z$, and
\begin{equation}
\L_3(Z[M],X[M]):= \sum_{i\mid M_{s,i}=1} \lambda_{3,i} \|Z_{\cdot,i}-X_{\cdot,i}\|^2_{\X_i^N}
\end{equation}
relaxes measurement/data constraints.
Minimizing over $f$ first in \eqref{eqhdewdgvgyvb} we obtain the following theorem.
\begin{Theorem}
 $f$, $Y$ and $Z$ are a minimizer of \eqref{eqhdewdgvgyvb} if and only if, for $f_e\in \cE_u$,
\begin{equation}\label{eqeersedgvgyvb}
f_e(\cdot)=K_e(\cdot, Z_{\cdot,a(e)}^\dagger) \big(K_e(Z_{\cdot,a(e)}^\dagger,Z_{\cdot,a(e)})+\lambda_{1,a(e)}^\dagger I\big)^{-1}Y_e^\dagger
\end{equation}
where $Y^\dagger$
and $Z^\dagger$ are a minimizer of
\begin{equation}\label{eqhdewdgvgyvb2b}
\begin{cases}
\minimize_{Z, Y} &\sum_{e\in \cE_u}Y_e^T \big(K_e(Z_{\cdot,a(e)},Z_{\cdot,a(e)})+\lambda_{1,a(e)} I\big)^{-1}Y_e\\&+
\sum_{e\in \cE/\cE_u} \lambda_{1,b(e)}\|f_e(Z_{\cdot,a(e)})-Y_{e}\|_{\X_{b(e)}^N}^2
\\&+\L_2( Y,Z)+\L_3(Z[M],X[M])\,,\\
\st &  Z_{{\cdot,j}}=\otimes_{i\leadsto j} Z_{\cdot,i}\,,\text{ for } j\in \V_\circ\,.
\end{cases}
\end{equation}
\end{Theorem}
Note that the proposed relaxation is  equivalent to the CGC solution \eqref{eqhdewdgvgyv} with a modification of the underlying computational graph corresponding to the addition of independent centered Gaussian random variable with covariance matrices/operators  (1) $I_{\X_{b(e)}}/\lambda_{1,b(e)} $ to the output of each function $f_e$, (2) $I_{\X_{j}}/\lambda_{2,j} $ to each node $j\in \V_\square^\iota$, (3) $I_{\X_{j}}/\lambda_{3,j} $ to each measurement of $X_{s,j}$.

\subsection{Feature maps and Artificial Neural Networks}
In the setting of Sec.~\ref{secrelcgc}, write $\psi_e$ and $\F_e$ for the feature map and space associated with the kernel $K_e$, i.e.
$\psi_e$ maps  $\X_{a(e)}$ to the space of bounded linear operators from $\X_{b(e)}$ to $\F_e$, and
$y^T K_e(z,z')y'= \<\psi_e(z) y,\psi_e(z') y'\>_{\F_e}$ for
$(x,x',y,y')\in \X^2_{a(e)} \times \X^2_{b(e)}$, which we write as $K_e(z,z')=\psi^T(z') \psi(z)$.
When the $\F_e$ are finite-dimensional then an alternative approach is to represent the unknown functions as $f_e=\psi^T_e \alpha_e$ where the $\alpha_e$ are  feature map coefficients determined by minimizing \eqref{eqhdewdgvgyvb} over the $\alpha_e$ and  $Z, Y$ with
\begin{equation}
\|f\|^2:=\sum_{e\in \cE_u}\|\alpha_e\|_{\F_e}^2
\end{equation}
and
\begin{equation}
\L_1(f,Z):=\sum_{e\in \cE} \lambda_{1,b(e)}\|\psi^T_e(Z_{\cdot,a(e)})\alpha_e-Y_{e}\|_{\X_{b(e)}^N}^2\,.
\end{equation}
If the feature maps $\psi_e$ themselves are parameterized by  hyperparameters $\theta_e$  then those hyperparameters can also be learned from data via MAP estimation (as in \cite{owhadi2020ideas}), Kernels Flows \cite{owhadi2019kernel, chen2021consistency} or MLE \cite{chen2021consistency}.
By interpreting ANNs as a particular case of parameterized feature map \cite{yoo2020deep, owhadi2020ideas} kernel learning techniques could be employed to solve CGC problems with ANNs. Note that  replacing unknown functions with ANNs is equivalent to replacing unknown functions with sub-graphs whose unknown functions are parameterized by the layers of the network (see Sec.~\ref{subsecekjhdedgh}). Furthermore  when applied to solving/learning PDEs (Sec.~\ref{seckjg67g} and \ref{seckjg67g}), and compared to prototypical ANN-based approaches (such as PINNs \cite{raissi2019physics}), the loss \eqref{eqhdewdgvgyvb} includes a RKHS regularization term on unknown functions and the relaxation of all composition steps, which are essential for the convergence and stability of the proposed method.
We will not develop further the ANN-based approach to CGC in this paper and refer to
 remarks \ref{rmkmot1} and \ref{rmkfgps} for motivations for its focus on GPs/kernel methods.

\subsection{MAP vs. empirical Bayes estimation}
The general CGC solutions presented in sec.~\ref{seccgcov} and \ref{secrelcgc} are based on MAP estimations of the underlying GPs.
To approximate the unknown functions of the graph with  empirical Bayes (EB) estimates of the underlying GPs, \eqref{eqhdeeedgyv} and \eqref{eqhdewdgvgyvb2b} must be updated as follows. In the non-relaxed case of Sec.~\ref{seccgcov}, EB approximates the unknown functions with \eqref{eqeersedgvgyv} where
where $(Y_e^\dagger)_{e\in \cE}$ and $Z^\dagger$ are a minimizer of
\begin{equation}\label{eqhdeeedgyveb}
\begin{cases}
\minimize_{Z, (Y_e)_{e\in \cE_u}}&\sum_{e\in \cE_u}Y_e^T K_e(Z_{\cdot,a(e)},Z_{\cdot,a(e)})^{-1}Y_e+\log \det K_e(Z_{\cdot,a(e)},Z_{\cdot,a(e)})
\\&+\sum_{s=1}^N \sum_{i\in \V_r}\|Z_{s,i}\|^2_{K_i}\\
\st& f_e(Z_{\cdot,a(e)})=Y_{\cdot,e} \,,\text{ for }e\in \cE/\cE_u\,,\\
{\rm and }& Z_{{\cdot,j}}=\sum_{e\leadsto j} Y_{\cdot,e}\,,\text{ for } j\in \V_\square^\iota \,, \\
{\rm and }& Z_{{\cdot,j}}=\otimes_{i\leadsto j} Z_{\cdot,i}\,,\text{ for } j\in \V_\circ\,,\\
{\rm and }& Z[M]=X[M]\,.
\end{cases}
\end{equation}
In the relaxed case of Sec.~\ref{secrelcgc}, EB approximates the unknown functions with \eqref{eqeersedgvgyvb}, where
 $Y^\dagger$
and $Z^\dagger$ are a minimizer of
\begin{equation}\label{eqhdewdgvgyvb2bedeb}
\begin{cases}
\minimize_{Z, Y} &\sum_{e\in \cE_u}Y_e^T \big(K_e(Z_{\cdot,a(e)},Z_{\cdot,a(e)})+\lambda_{1,a(e)} I\big)^{-1}Y_e
\\&+\log \det K_e(Z_{\cdot,a(e)},Z_{\cdot,a(e)}+\lambda_{1,a(e)} I)
\\&+
\sum_{e\in \cE/\cE_u} \lambda_{1,b(e)}\|f_e(Z_{\cdot,a(e)})-Y_{e}\|_{\X_{b(e)}^N}^2+\L_2( Y,Z)+\L_3(Z[M],X[M])\,,\\
\st &  Z_{{\cdot,j}}=\otimes_{i\leadsto j} Z_{\cdot,i}\,,\text{ for } j\in \V_\circ\,.
\end{cases}
\end{equation}

We refer to \cite{dunlop2020hyperparameter} for detailed comparisons between MAP and EB estimators.
In particular, we note that (1) MAP estimators are generally not invariant under change of parameterization, (2) EB estimators are invariant under change of parameterization. Furthermore, MAP estimators may, in some situations, lack the consistency of EB estimators. One of those situations is described in \cite{darcy2022learningcgc} which employs the \eqref{eqhdeeedgyveb} and \eqref{eqhdewdgvgyvb2bedeb} variational formulae to estimate (in the CGC framework)  the drift and diffusion parameters of an SDE from sparse observations of its trajectory.
In all other situations (where MAP estimators are consistent, e.g., mode decomposition, idea registration), MAP estimators are generally preferred due to their lower complexity.

\section{Further examples}\label{seckjg67g00}
We will now illustrate the flexibility and scope of the CGC framework
through examples. These examples include the seamless CGC representation of known methods  and the discovery of new ones.

\subsection{Solving nonlinear PDEs}\label{seckjg67g}
We will now show that GP approach to solving nonlinear PDEs, introduced in \cite{chen2021solving}, can naturally be represented as a CGC solution. As in \cite{chen2021solving} the nonlinear PDE can be arbitrary but, for ease of presentation, we will only consider the  following prototypical nonlinear elliptic PDE \cite[Sec.~1.1]{chen2021solving}
\begin{equation}
  \label{elliptic-proto-PDE}
  \left\{
    \begin{aligned}
      -\Delta u(x) + \tau\big(u(x)\big) &= f(x), &&\forall
      x \in \Omega\,, \\
      u(x) & = g(x), &&\forall
      x \in \partial\Omega\, ,
    \end{aligned}
    \right.
\end{equation}
where $\tau$ is a  nonlinearity (e.g., $\tau(u)=u^3$) and $\Omega\subset \R^d$ and $\tau, f, g$ are continuous and  such that \eqref{elliptic-proto-PDE} admits a unique strong solution in $H^s(\Omega)$ with $s>d/2+2$ (so that $u$ has continuous $C^2$ derivatives).
To approximate the solution of \eqref{elliptic-proto-PDE}  as a CGC problem, let
 $X_1,\ldots,X_M$ and $X_{M+1},\ldots,X_N$ be a finite number of collocation points on $\Omega$ and $\partial \Omega$. Then, as illustrated in the following diagram, consider the problem of
 approximating the unknown function $u$ given  $u(X_i)=g(X_i)$ for $i=M+1,\ldots,N$ and  $(-\Delta u+\tau(u))(X_i)=f(X_i)$ for $i=1,\ldots,M$.\\
 \centerline{
\begin{tikzpicture}[->,>=stealth',shorten >=1pt,auto,node distance=3cm,
                    thick,main node/.style={rectangle,draw,font=\sffamily\Large\bfseries}]

\node[main node] (1) {$x$};
\node (2) [main node, right of=1, node distance=4cm] {$(u(x),-\Delta u(x))$};
\node (3) [main node, right of=2, node distance=7cm] {$-\Delta u(x)+\tau(u(x))$};
\node (4) [main node, above of=2, node distance=2cm] {$u(x)$};

\path[every node/.style={font=\sffamily\Large\bfseries},red]
    (1) edge node [above,red] {$u$} (2);

\path[every node/.style={font=\sffamily\Large\bfseries}]
    (2) edge node [above,style={font=\sffamily\Large\bfseries} ] {$z_2+\tau(z_1)$} (3);

%\path[every node/.style={font=\sffamily\Large\bfseries}]
%    (1) edge  [bend right,dashed ] node[above ] {$f$} (3);

  \tikzstyle{every to}=[draw,dashed]
\draw[dashed] (1) to[out=-45,in=-135,looseness=0.2,style={font=\sffamily\Large\bfseries,dashed}] node[below] {$(X_i,f(X_i))_{1\leq i \leq M}$} (3);

\path[every node/.style={font=\sffamily\Large\bfseries}]
    (2) edge node [right,style={font=\sffamily\Large\bfseries} ] {$z_1$} (4);

\path[every node/.style={font=\sffamily\small\bfseries}]
    (1) edge  [dashed ] node[sloped, anchor=center, above] {\quad$(X_i,g(X_i))_{M+1\leq i \leq N}$} (4);

\end{tikzpicture}
}
Note that we are using the setting of Sec.~\ref{secgenvar}.
Replacing $u$ with a centered GP $\xi \sim \cN(0,K)$ with kernel  $K\,:\bar{\Omega}\times \bar{\Omega}\rightarrow \R$ (chosen so that  $\H_K\subset C^2(\Omega)\cap C(\bar{\Omega})$), we recover the solution introduced in \cite{chen2021solving}. As detailed in  \cite{chen2021solving}, this solution  is to approximate $u$ with the MAP estimator of $\xi$ given the data (given $\xi(X_i)=g(X_i)$ for $i=M+1,\ldots,N$ and  $-(\Delta \xi+\tau(\xi))(X_i)=f(X_i)$ for $i=1,\ldots,N$), which  can  be computed as a minimizer of
   \begin{equation}\label{running-example-two-level-optimization-problem}
     \left\{
     \begin{aligned}
       & \minimize_{z^{(1)} \in \R^N, z^{(2)} \in \R^{M}} \: \left\{
       \begin{aligned}
          & \minimize_{u} ~\|u\|\\
          &\st \quad  u(X_i)=z^{(1)}_i \text{ and } -\Delta u(x_i)=z^{(2)}_i,\text{ for } i=1,\ldots,M,
       \end{aligned} \right.\\
  &\st\quad z^{(2)}_i +\tau(z^{(1)}_i)=f(X_i), \hspace{28ex} \text{for } i=1,\ldots,M\,, \\
  &     \hspace{6ex}   z^{(1)}_i=g(X_i),  \hspace{37.5ex} \text{for } i=M+1,\ldots, N\,,
\end{aligned}
\right.
\end{equation}
where the intermediate variables $z^{(1)}_i:=u(X_i)$ and $z^{(2)}_i:=\Delta u(X_i)$ appear as nodes in the computational graph.
After reduction, \eqref{running-example-two-level-optimization-problem} can then be solved using a variant of the Gauss-Newton algorithm  \cite{chen2021solving}, which implies that the complexity of the method inherits that of the state-of-the-art solvers for inverting/compressing dense kernel matrices (e.g., in $N\log^{2d}N$ complexity with a variant of \cite{schafer2021sparse}).
Furthermore, \cite{chen2021solving} also shows that if the kernel $K$ is adapted in the sense that $u\in \H_K$ and $\H_K \subseteq H^s(\Omega)$   for some $s >2+d/2$, then the proposed approach is guaranteed to converge (pointwise and in $\H^t(\Omega)$ for $t<s$) as the fill distance between collocation points goes to zero \cite[Thm.~1.2]{chen2021solving}.

\begin{Remark}\label{rmkfgps}
Current learning methods for scientific computing can be divided into two main categories: (1)
methods based on variants of artificial neural networks (ANNs) with Physics Informed Neural Networks as a prototypical example \cite{raissi2019physics}; and (2) methods based on kernels and Gaussian Processes (GPs) \cite{williams1996gaussian,scholkopf2018learning,  OwhScoSchNotAMS2019, Hennig2015, raissi2017inferring, cockayne2019bayesian} with gamblets \cite{Owhadi:2014, owhadi2017multigrid, owhadi2019operator} as a prototypical example. Methods of type (2) were essentially limited to linear/quasi-linear/time-dependent PDEs, and they have been recently generalized to arbitrary nonlinear PDEs \cite{chen2021solving}. As discussed in \cite{chen2021solving},
methods of type (2) hold potential for considerable advantages over methods of type (1), in terms of theoretical
analysis, numerical implementation, regularization, guaranteed convergence, automatization, interpretability, and higher-level reasoning.
Furthermore, by interpreting ANNs as data-dependent feature maps, methods
of type (1) can be analyzed, generalized, and regularized as methods of type (2) with data adapted kernels  \cite{owhadi2020ideas, belkin2021fit}.
\end{Remark}

 \begin{figure}[h]
	\begin{center}
			\includegraphics[width= \textwidth]{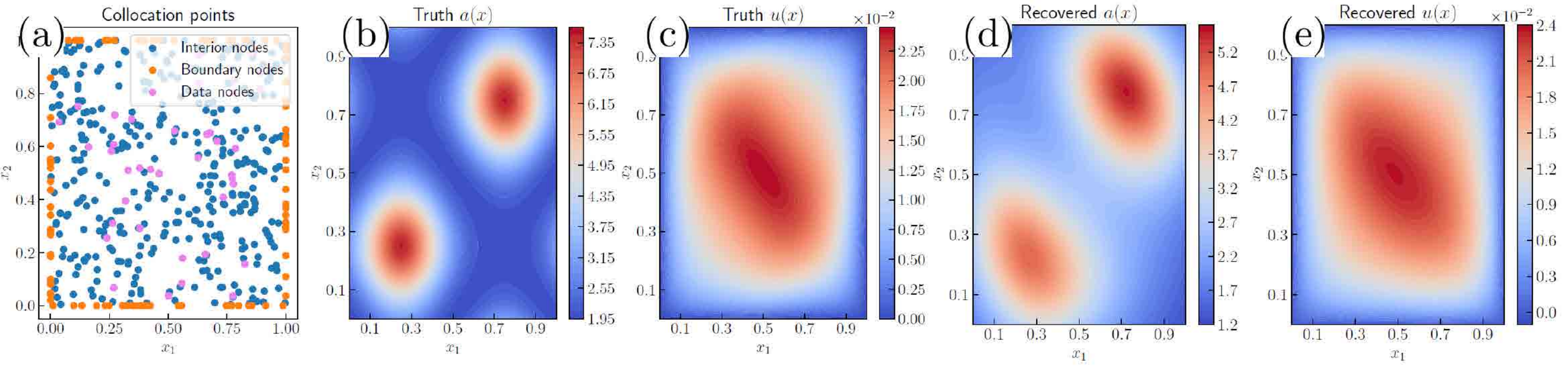}
		\caption{\cite[Fig.~5]{chen2021solving}. (a) $u$ is observed at $40$ data points (colored in magenta) and the PDE is enforced at $400$ (interior and boundary) collocation points;  (b) true $a$;  (c) true $u$; (d) recovered $a$; (e) recovered $u$. The accuracy between true and recovered is a function of the number data points which is small ($40$) in this illustration.}\label{figdf}
	\end{center}
\end{figure}

\subsection{Learning  PDEs}\label{seckjg67gb}
In our next example we will show that GP approach to learning PDEs, introduced in \cite{chen2021solving}, can also be formulated as a CGC solution. The  PDE can be nonlinear and arbitrary but, for ease of presentation, but we will only consider the   prototypical  inverse problem \cite[Sec.~4]{chen2021solving} of approximating the coefficient $a$ and the solution $u$ of the following PDE ($\Omega = (0,1)^2$)
  \begin{equation}\label{eqwkjgdjhedjh}
    \left\{
    \begin{aligned}
      - \operatorname{div} \left(   \exp(a) \nabla u \right) (x)  & = f (x),  && x \in \Omega\, , \\
          u(x) & = g,   && x \in \partial \Omega\, .
  \end{aligned}
  \right.
\end{equation}
given the observation of the value $Y_i$ of $u$ at a finite number of collocation points $X_i$.
Consider the  following CGC diagram above  and recover $a$, $u$ from the measurements  $u(X_i)=Y_i$  at a finite number of data points $X_i$ (colored in magenta in Fig.~\ref{figdf}.(a)) and of the values of  $- \operatorname{div} \left(   \exp(a) \nabla u \right)(X_s)=f(X_s)$ and $u(X_j)=g(X_j)$ at a finite number of collocation points $X_s\in \Omega$ and $X_j\in \partial \Omega$.\\
\centerline{\scalebox{0.95}{
\begin{tikzpicture}[->,>=stealth',shorten >=1pt,auto,node distance=3cm,
                    thick,main node/.style={rectangle,draw,font=\sffamily\large\bfseries}]
\useasboundingbox (0,0) rectangle (12,3.5);
\node[main node] (1) {$x$};
\node (2) [main node, right of=1, node distance=4cm] {$(u(x),\nabla u(x),\Delta u(x))$};
\node (4) [main node, above of=2, node distance=2cm] {$u(x)$};
\node (5) [main node, below of=1, node distance=2cm] {$(a(x),\nabla a(x))$};
\node (6) [main node, circle, right of=5, node distance=4cm] {};
\node (3) [main node, right of=6, node distance=7cm,font=\sffamily\bfseries] {$-e^{a(x)} (\Delta u(x)+\nabla a(x)\nabla u(x))$};

\path[every node/.style={font=\sffamily\large\bfseries},red]
    (1) edge node [above,red] {$u$} (2)
    (1) edge node [left,red] {$a$} (5);

\path[every node/.style={font=\sffamily\large\bfseries}]
    (2) edge node [right,style={font=\sffamily\large\bfseries} ] {$(z_1,z_2,z_3)$} (6);

\path[every node/.style={font=\sffamily\small}]
    (1) edge  [bend left,dashed ] node[sloped, anchor=center, above ] {$(X_s,Y_s)$} (4);

\tikzstyle{every to}=[draw,dashed]
\draw[dashed] (1) to[out=90,in=90,looseness=1,style={font=\sffamily\small\bfseries,dashed}] node[sloped, anchor=center, above] {$(X_i,f(X_i))$} (3);

%\path[every node/.style={font=\sffamily\large}]
%    (1) edge  [ bend left,dashed ] node[above ] {$f$} (3);

\path[every node/.style={font=\sffamily\large\bfseries}]
    (2) edge node [right,style={font=\sffamily\large\bfseries} ] {$z_1$} (4);

\path[every node/.style={font=\sffamily\small\bfseries}]
    (1) edge  [dashed ] node[sloped, anchor=center, above ] {$(X_j,g(X_j))$} (4);

\path[every node/.style={font=\sffamily\large\bfseries}]
    (5) edge node [below,style={font=\sffamily\large\bfseries} ] {$(z_4,z_5)$} (6);

\path[every node/.style={font=\sffamily\large\bfseries}]
    (6) edge node [below,style={font=\sffamily\large\bfseries} ] {$-e^{z_4} (z_3+z_5 z_2)$} (3);

\end{tikzpicture}}}
\vskip 28mm
The  GP solution to this CGC problem  recovers that of \cite[Sec.~4]{chen2021solving} illustrated in
Fig.~\ref{figdf}. Given two  kernels $\Gamma$ and $K$ (such that $\H_\Gamma$ and $\H_K$ contain $C^1$ and $C^2$ functions), this solution is to identify $a$ and $u$ as minimizers of
{\small
\begin{equation}\label{inverse-problem-optimization-form-standard}
      \begin{cases}
        & \minimize_{u,a,z^{(1)}, \ldots, z^{(5)} }
        \quad   \| u\|_K^2  +  \| a\|_\Gamma^2\\
  &\st    \Delta u(X_i)=(z^{(3)}_i,\, \nabla u(X_i)= z^{(2)}_i,\, a(X_i)=z^{(4)}_i,\,\nabla a(X_i)=z^{(5)}_i,   \text{ for } i=1,\ldots,M\,, \\
  & \hspace{6ex} u(X_j)=z^{(1)}, \text{ for } j=M+1,\ldots, N\,,\\
  & \hspace{6ex} u(X_s)=Y_s  \text{ for } s=N+1,\ldots, N+D\,,\\
   & \hspace{6ex} -e^{z^{(4)}}_i (z^{(3)}_i+z^{(5)}_i z^{(2)}_i)(X_i)=f(X_i),  \text{ for } i=1,\ldots,M\,, \\
  & \hspace{6ex} z^{(1)}=g(X_j), \text{ for } j=M+1,\ldots, N\,,\\
  & \hspace{6ex} z^{(1)}=Y_s  \text{ for } s=N+1,\ldots, N+D\,.
      \end{cases}
\end{equation}
}

\subsection{ResNet and idea registration}\label{subsecekjhdedgh}
Training ANNs can naturally be formulated as a particular case of CGC.
In this example, taken from \cite{owhadi2020ideas}, we consider the problem of regressing the data $(X_i,Y_i)_{i=1}^N$ with a function of the form
 \begin{equation}\label{eqkeljbekjbdedbh}
f\circ \phi_L\text{ where } \phi_L=(I+v_L)\circ \cdots \circ (I+v_1)\,,
  \end{equation}
  where $f,v_1,\ldots,v_L$ are unknown functions and $I$ is the identity function.
  Introducing the intermediate variables $q_1=x+v_1(x)$, $q_2=q_1+ v_2(q_1)$ and $q_3=q_2+ v_3(q_2)$, this nonlinear regression problem can be formulated as the CGC problem illustrated by the following graph for  $L=3$.\\
\centerline{
\begin{tikzpicture}[->,>=stealth',shorten >=1pt,auto,node distance=2.5cm,
                    thick,main node/.style={rectangle,draw,font=\sffamily\Large\bfseries}]

\node[main node] (1) {$x$};
\node[main node] (2) [right of=1]  {$q_2$};
\node[main node] (3) [right of=2] {$q_3$};
\node[main node] (4) [right of=3] {$q_4$};
\node[main node] (5) [right of=4] {$y$};
\node[main node] (6) [above of=5,blue,node distance=1.5cm] {$w^4$};

\path[every node/.style={font=\sffamily\Large\bfseries}]
    (1) edge node [below ] {} (2)
    (1) edge  [bend left,red] node[above ] {$v_1$} (2);

\path[every node/.style={font=\sffamily\Large\bfseries}]
    (2) edge node [below ] {} (3)
    (2) edge  [bend left,red] node[above ] {$v_2$} (3);

\path[every node/.style={font=\sffamily\Large\bfseries}]
    (3) edge node [below ] {} (4)
    (3) edge  [bend left,red] node[above ] {$v_3$} (4);

\path[every node/.style={font=\sffamily\Large\bfseries},red]
    (4) edge node [above ] {$f$} (5);

\path[every node/.style={font=\sffamily\Large\bfseries}]
    (6) edge node [right ] {} (5);

%\path[every node/.style={font=\sffamily\Large}]
%    (1) edge  [bend right,dashed ] node[above ] {data} (5);

\tikzstyle{every to}=[draw,dashed]
\draw[dashed] (1) to[out=-90,in=-90,looseness=0.3,style={font=\sffamily\Large,dashed}] node[above] {$(X,Y)$} (5);

\end{tikzpicture}}
Given  kernels $\Gamma$ and $K$, $f$ and the $v_i$ can be identified as minimizers of
\begin{equation}\label{eqhjejhdejhddh}
\min_{f,v_1,\ldots,v_L}\frac{\nu L}{2}\sum_{s=1}^L \|v_s\|_{\Gamma}^2+ \lambda \|f\|_{K}^2+\|f\circ \phi_L (X)-Y\|_{\Y^N}^2\,,
\end{equation}
over $f\in \H_K$ and $v_i\in \H_\Gamma$.
As noted in \cite{owhadi2020ideas}, if $\Gamma$ and $K$ are affine scalar operator valued kernels (of the form $\Gamma(x,x')=(\bf(a)(x)\ba(x')+1)I_\X$ and $K(x,x')=(\bf(a)(x)\ba(x')+1)I_\Y$ where $\ba$ is an activation function defined as an elementwise nonlinearity), then $f\circ \phi_L$ has the structure of one ResNet block
\cite{he2016deep}  and minimizing \eqref{eqhjejhdejhddh} is equivalent to training that network with $L_2$ regularization on weights and biases.

Letting $q^1:=X$, and introducing the intermediate variables $q^{s+1}=q^s+v_s(q^s)$ for $1\leq s \leq L$ we obtain that
 $(f,v_1,\ldots,v_L)$ is  a minimizer of \eqref{eqhjejhdejhddh} if and only if
\begin{equation}\label{eqkjdkedkjnd}
v_s(x)= \wK( x, q^s) \wK(q^s,q^s)^{-1} (q^{s+1}-q^s) \text{ for } x\in \X, s\in \{1,\ldots,L\}\,,
\end{equation}
and
\begin{equation}\label{eqkeewedkjnd}
f(x)= K( x, q^{L+1}) \big(K(q^{L+1},q^{L+1}+\lambda^{-1} I\big)^{-1} Y
\end{equation}
where  $q^1,\ldots,q^{L+1} \in \X^N$ is a minimizer of (write $\Delta t:=1/L$) the discrete least action principle
\begin{equation}\label{eqlsedhjd}
\begin{cases}
\text{Minimize } &\frac{\nu}{2}  \sum_{s=1}^L  (\frac{q^{s+1}-q^s}{\Delta t})^T \wK(q^s,q^s)^{-1} (\frac{q^{s+1}-q^s}{\Delta t})\, \Delta t+
\lambda Y^T \big(K(q^{L+1},q^{L+1}+\lambda^{-1} I\big)^{-1} Y \\
\text{over } &q^2,\ldots,q^{L+1} \in \X^N
\text{ with }q^1=X\,.
\end{cases}
\end{equation}
Furthermore, as $L\rightarrow \infty$, \cite{owhadi2020ideas} establishes the convergence of the trajectory formed by the $q^s$ towards a continuous least action principle and deduces the convergence of $f\circ \phi_L$ towards $f\circ \phi^v(\cdot,1)$ where $\phi^v(\cdot,t)$ is the flow map of the vector field $v$ ($\phi(x,0)=x$ and $\dot{\phi}(x,t)=v(\phi(x,t),t)$), and $(f,v)$ are identified  as a minimizer of the following variational problem
\begin{equation}\label{eqhjeddedjhddh}
\min_{f,v}\frac{\nu }{2}\int_0^1 \|v(x,t)\|_{\Gamma}^2+ \lambda \|f\|_{K}^2+\|f\circ \phi(X,1)-Y\|_{\Y^N}^2\,.
\end{equation}
\eqref{eqhjeddedjhddh}, which
has the structure of variational formulations used in computational anatomy \cite{grenander1998computational}, image registration \cite{brown1992survey} and shape analysis \cite{younes2010shapes}, can be seen as a generalization of  image registration problem with images  replaced by high dimensional shapes/forms (ideas).

\subsection{Regularized ResNet}\label{secrefkwhwd88d}

The regressor $f\circ \phi_L$ obtained in \eqref{subsecekjhdedgh} is not continuous with respect to the data $(X,Y)$ and unstable with respect to the input variable $x$ \cite{owhadi2020ideas}. A natural and rigorous approach to its regularization (which ensures the continuity and stability of the regressor), presented in \cite{owhadi2020ideas}, is to add random (noise) variables $w_i$ to intermediate variables, as
illustrated in the following diagram for $L=3$.\\
\centerline{
\begin{tikzpicture}[->,>=stealth',shorten >=1pt,auto,node distance=2.5cm,
                    thick,main node/.style={rectangle,draw,font=\sffamily\Large\bfseries}]

\node[main node] (1) {$x$};
\node[main node] (2) [right of=1]  {$q_2$};
\node[main node] (3) [right of=2] {$q_3$};
\node[main node] (4) [right of=3] {$q_4$};
\node[main node] (5) [right of=4] {$y$};
\node[main node] (6) [above of=2,blue,node distance=1.5cm] {$w_1$};
\node[main node] (7) [above of=3,blue,node distance=1.5cm] {$w_2$};
\node[main node] (8) [above of=4,blue,node distance=1.5cm] {$w_3$};
\node[main node] (9) [above of=5,blue,node distance=1.5cm] {$w_4$};

\path[every node/.style={font=\sffamily\Large\bfseries}]
    (1) edge node [below ] {} (2)
    (1) edge  [bend left,red] node[above ] {$v_1$} (2);

\path[every node/.style={font=\sffamily\Large\bfseries}]
    (2) edge node [below ] {} (3)
    (2) edge  [bend left,red] node[above ] {$v_2$} (3);

\path[every node/.style={font=\sffamily\Large\bfseries}]
    (3) edge node [below ] {} (4)
    (3) edge  [bend left,red] node[above ] {$v_3$} (4);

\path[every node/.style={font=\sffamily\Large\bfseries},red]
    (4) edge node [below ] {$f$} (5);

\path[every node/.style={font=\sffamily\Large\bfseries}]
    (6) edge node [right ] {} (2);

\path[every node/.style={font=\sffamily\Large\bfseries}]
    (7) edge node [right ] {} (3);

\path[every node/.style={font=\sffamily\Large\bfseries}]
    (8) edge node [right ] {} (4);

\path[every node/.style={font=\sffamily\Large\bfseries}]
    (9) edge node [right ] {} (5);

%\path[every node/.style={font=\sffamily\Large}]
%    (1) edge  [bend right,dashed ] node[above ] {data} (5);

\tikzstyle{every to}=[draw,dashed]
\draw[dashed] (1) to[out=-90,in=-90,looseness=0.3,style={font=\sffamily\Large,dashed}] node[above] {$(X,Y)$} (5);

\end{tikzpicture}
}
With this modification $(f,v_1,\ldots,v_L)$ is then identified as a minimizers of
 \begin{equation}\label{eqlktddsyeytdsedhjdreg}
\begin{cases}
\text{Minimize } &\frac{\nu}{2}\,L\sum_{s=1}^L \big(\|v_s\|_{\Gamma}^2+ \frac{1}{r} \|q^{s+1}-(I+v_s)(q^s)\|_{\X^N}^2\big)
\\&+\lambda\,\|f\|_{\H}^2+ \|f(q^{L+1})-Y\|_{\Y^N}^2 \big)\\
\text{over }&v_1,\ldots,v_L \in \H_\Gamma,\, f\in \H_K,\, q^1,\ldots,q^{L+1} \in \X^N,\, q^1=X,\,,
\end{cases}
\end{equation}
Furthermore, $(v_1,\ldots,v_L ,f)$,  is  a minimizer of \eqref{eqlktddsyeytdsedhjdreg} if and only if $f=$\eqref{eqkeewedkjnd} and
\begin{equation}\label{eqkjdkedkjndreg}
v_s(x)= \wK( x, q^s) (\wK(q^s,q^s)+r I)^{-1} (q^{s+1}-q^s) \text{ for } x\in \X, s\in \{1,\ldots,L\}\,,
\end{equation}
where $q^1,\ldots,q^{L+1} \in \X^N$ is a minimizer of (write $\Delta t:=1/L$)
\begin{equation}\label{eqlsedhjdreg}
\begin{cases}
\text{Minimize } &\frac{\nu}{2}  \sum_{s=1}^L  (\frac{q^{s+1}-q^s}{\Delta t})^T (\wK(q^s,q^s)+r I)^{-1} (\frac{q^{s+1}-q^s}{\Delta t})\, \Delta t\\&+
\lambda Y^T \big(K(q^{L+1},q^{L+1}+\lambda^{-1} I\big)^{-1} Y\\
\text{over } &q^2,\ldots,q^{L+1} \in \X^N
\text{ with }q^1=X\,,
\end{cases}
\end{equation}
The regularization process illustrated here is generic and, as discussed in Sec.~\ref{secrelcgc}, can be extended to all CGC problems by adding noise variables (acting as nuggets in the setting of kriging) to all intermediate variables.

\subsection{Dimension reduction}\label{subseckhey}
We will now present several variants of dimension reduction methods identified as solutions of CGC problems.
\subsubsection{Kernel variant of autoencoders}\label{subseckhey01}
Consider the following diagram.\\
\centerline{
\begin{tikzpicture}[->,>=stealth',shorten >=1pt,auto,node distance=3cm,
                    thick,main node/.style={rectangle,draw,font=\sffamily\Large\bfseries}]

\node[main node] (1) {$x$};
\node[main node] (2) [right of=1] {$z$};
\node[main node] (3) [right of=2] {$y$};
\node[main node] (4) [above of=2,blue, node distance=1cm] {$w_1$};
\node[main node] (5) [above of=3,blue, node distance=1cm] {$w_2$};

\path[every node/.style={font=\sffamily\Large\bfseries},red]
    (1) edge node [above ] {$g$} (2);

\path[every node/.style={font=\sffamily\Large\bfseries},red]
    (2) edge node [above ] {$f$} (3);

\path[every node/.style={font=\sffamily\Large\bfseries}]
    (4) edge node [above ] {} (2);

\path[every node/.style={font=\sffamily\Large\bfseries}]
    (5) edge node [right ] {} (3);

\path[every node/.style={font=\sffamily\Large}]
    (1) edge  [bend right,dashed ] node[below ] {$(X,X)$} (3);

\end{tikzpicture}}
In that diagram we consider the problem of recovering two unknown functions $f$ and $g$ given the data
$(X_i=f(g(X_i)+W_{1,i})+W_{2,i})_{i=1,\ldots,N}$, where the  $W_{1,i}$ and $W_{2,i}$ are i.i.d. $\cN(0,\sigma_1^2)$ and $\cN(0,\sigma_2^2)$ random variables. Writing $\dim(x)$ for the dimension of the variable $x$, we select $\dim(x)=\dim(y)>\dim(z)$, so that the intermediate $z$ variables act as reduced/coarse/latent variables. Writing $\X$ for the domain of $x$ and $y$ and $\Z$ for the domain of $z$,
given two kernels $\Gamma\,:\, \X\times \X \rightarrow \L(\Z)$ and $K\,: \Z\times \Z\rightarrow \L(\X)$, the proposed GP approach is to   consider the GPs $\xi_1 \sim \cN(0,\Gamma)$ and $\xi_2 \sim \cN(0,K)$ and approximate $(g,f)$ with a MAP estimator of $(\xi_1,\xi_2)$ given
$(X=\xi_2(\xi_1(X)+W_{1})+W_{2})$. Introducing $Z=g(X)+W_1$ as an intermediate variable, these MAP estimators can be identified as  minimizers of\,,
\begin{equation}\label{eqhgvygdeee}
\minimize_{f\in \H_K, g\in \H_{\Gamma}, Z\in \Z^N} \|g\|_{\Gamma}^2+\|f\|_{K}^2+\frac{1}{\sigma^2_1}\|g(X)-Z\|^2_{\Z^N}+\frac{1}{\sigma^2_2}\|f(Z)-X\|^2_{\X^N}\,.
\end{equation}
Following \cite{owhadi2020ideas} we minimize over $f$ and $g$ first and obtain that
\begin{equation}\label{eqagteydhje33bdjehor}
f^\dagger(x)=K(x,Z^\dagger)\big(K(Z^\dagger,Z^\dagger)+\sigma_2^2 I\big)^{-1} X\,,
\end{equation}
and
\begin{equation}\label{eqagwtydhje33bdjehor}
g^\dagger(x)=\Gamma(x,X)\big(\Gamma(X,X)+\sigma_1^2 I\big)^{-1} Z^\dagger\,,
\end{equation}
where $Z^\dagger$ is a minimizer of
\begin{equation}\label{eqhgdqerdee}
\minimize_{ Z\in \Z^N}Z^T \big(\Gamma(X,X)+\sigma_1^2 I \big)^{-1} Z+X^T \big(K(Z,Z)+\sigma_2^2 I\big)^{-1} X\,.
\end{equation}
Note that if $f$ and $g$ are replaced by ANNs, then $f\circ g$ has the structure of an autoencoder  \cite{baldi2012autoencoders}.
Furthermore, if one selects $K$ and $\Gamma$ to be linear kernels (i.e., having linear feature maps) and $\sigma_1=0$, then one rediscovers  SVD/PCA \cite{golub1971singular} as the solution of the proposed CGC problem.

\subsubsection{Kernel PCA}\label{subseckhey02}
In the following example we reformulate Kernel PCA \cite{mika1998kernel} as the solution of the CGC problem illustrated by the following diagram.\\
\centerline{
\begin{tikzpicture}[->,>=stealth',shorten >=1pt,auto,node distance=3cm,
                    thick,main node/.style={rectangle,draw,font=\sffamily\Large\bfseries}]

\node[main node] (1) {$x$};
\node[main node] (2) [right of=1] {$x'$};
\node[main node] (3) [right of=2] {$z$};
\node[main node] (4) [right of=3] {$y$};
\node[main node] (5) [above of=4,blue, node distance=1cm] {$w$};

\path[every node/.style={font=\sffamily\Large\bfseries}]
    (1) edge node [above ] {$\varphi$} (2);

\path[every node/.style={font=\sffamily\Large\bfseries},red]
    (2) edge node [above ] {$g$} (3);

\path[every node/.style={font=\sffamily\Large\bfseries},red]
    (3) edge node [above ] {$f$} (4);

\path[every node/.style={font=\sffamily\Large\bfseries}]
    (5) edge node [right ] {} (4);

%\path[every node/.style={font=\sffamily\Large}]
%    (1) edge  [bend right,dashed ] node[below ] {$(X,\varphi(X))$} (4);

  \tikzstyle{every to}=[draw,dashed]
\draw[dashed] (1) to[out=-45,in=-135,looseness=0.3,style={font=\sffamily\Large\bfseries,dashed}] node[below] {$(X,\varphi(X))$} (4);

\end{tikzpicture}
}
This is done by interpreting $\varphi$ as the feature map of a given kernel,    selecting $\dim(x')=\dim(y)>\dim(z)$, and recovering $f$ and $g$ with linear kernels.

\subsubsection{Nonlinear active subspace}\label{subseckhey03}

We now present a  generalization of active subspace methods \cite{constantine2014active}  to nonlinear spaces. Consider the following diagram.\\
\centerline{
\begin{tikzpicture}[->,>=stealth',shorten >=1pt,auto,node distance=3cm,
                    thick,main node/.style={rectangle,draw,font=\sffamily\Large\bfseries}]

\node[main node] (1) {$x$};
\node[main node] (2) [right of=1] {$z$};
\node[main node] (3) [right of=2] {$y$};
\node[main node] (4) [above of=2,blue, node distance=1cm] {$w_1$};
\node[main node] (5) [above of=3,blue, node distance=1cm] {$w_2$};

\path[every node/.style={font=\sffamily\Large\bfseries},red]
    (1) edge node [above ] {$g$} (2);

\path[every node/.style={font=\sffamily\Large\bfseries},red]
    (2) edge node [above ] {$f$} (3);

\path[every node/.style={font=\sffamily\Large\bfseries}]
    (4) edge node [above ] {} (2);

\path[every node/.style={font=\sffamily\Large\bfseries}]
    (5) edge node [right ] {} (3);

\path[every node/.style={font=\sffamily\Large}]
    (1) edge  [bend right,dashed ] node[below ] {$(X,F(X))$} (3);

\end{tikzpicture}}
In that diagram $F\,:\X\rightarrow \Y$ is a computable function and we seek to reduce its dimension by selecting $\dim(z)<\min\big(\dim(x),\dim(y)\big)$ and approximating $F$ with $f\circ g$. Given $(X,F(X))$, where $X$ is a set of points in the input space, and given two kernels $K$ and $\Gamma$, the proposed CGC solution is to obtain $(f,g)$ as minimizer of
\begin{equation}\label{eqhgvygdeeeV}
\minimize_{f\in \H_K, g\in \H_{\Gamma}, Z\in \Z^N} \|g\|_{\Gamma}^2+\|f\|_{K}^2+\frac{1}{\sigma^2_1}\|g(X)-Z\|^2_{\Z^N}+\frac{1}{\sigma^2_2}\|f(Z)-F(X)\|^2_{\X^N}\,.
\end{equation}
which is a simple variant of \eqref{eqhgvygdeee}, whose minimizers are of the form
\begin{equation}\label{eqagteydhje33bdjehorV}
f^\dagger(x)=K(x,Z^\dagger)\big(K(Z^\dagger,Z^\dagger)+\sigma_2^2 I\big)^{-1} F(X)\,,
\end{equation}
and
\begin{equation}\label{eqagwtydhje33bdjehorV}
g^\dagger(x)=\Gamma(x,X)\big(\Gamma(X,X)+\sigma_1^2 I\big)^{-1} Z^\dagger\,,
\end{equation}
where $Z^\dagger$ is a minimizer of
\begin{equation}\label{eqhgdqerdeeb}
\minimize_{ Z\in \Z^N}Z^T \big(\Gamma(X,X)+\sigma_1^2 I \big)^{-1} Z+F(X)^T \big(K(Z,Z)+\sigma_2^2 I\big)^{-1} F(X)\,.
\end{equation}
Note that if $f$ and $g$ are replaced by linear maps and if $F$ is linear then (for $\sigma_1=0$)   one rediscovers  the SVD approximation of $F$.

\subsubsection{Nonlinear PCA variant}\label{subseckhey04}
We now present a dimension reduction variant obtaining by combining the diagrams of subsections \ref{subseckhey} and \ref{secrefkwhwd88d}.\\
\centerline{\scalebox{0.7}{
\begin{tikzpicture}[->,>=stealth',shorten >=1pt,auto,node distance=2.5cm,
                    thick,main node/.style={rectangle,draw,font=\sffamily\Large\bfseries}]

\node[main node] (1) {$x$};
\node[main node] (2) [right of=1]  {$q_2$};
\node[main node] (3) [right of=2] {$q_3$};
\node[main node] (4) [right of=3] {$q_4$};
\node[main node] (5) [right of=4] {$z$};
\node[main node] (6) [above of=2,blue,node distance=1.5cm] {$w_1$};
\node[main node] (7) [above of=3,blue,node distance=1.5cm] {$w_2$};
\node[main node] (8) [above of=4,blue,node distance=1.5cm] {$w_3$};
\node[main node] (9) [above of=5,blue,node distance=1.5cm] {$w_4$};
\node[main node] (10) [right of=5] {$q_4'$};
\node[main node] (11) [right of=10] {$q_3'$};
\node[main node] (12) [right of=11] {$q_2'$};
\node[main node] (13) [right of=12] {$y$};
%\node[main node] (14) [right of=13] {$y$};
\node[main node] (15) [above of=10,blue,node distance=1.5cm] {$w_4'$};
\node[main node] (16) [above of=11,blue,node distance=1.5cm] {$w_3'$};
\node[main node] (17) [above of=12,blue,node distance=1.5cm] {$w_2'$};
\node[main node] (18) [above of=13,blue,node distance=1.5cm] {$w_1'$};

\path[every node/.style={font=\sffamily\Large\bfseries}]
    (1) edge node [below ] {} (2)
    (1) edge  [bend left,red] node[above ] {$v_1$} (2);

\path[every node/.style={font=\sffamily\Large\bfseries}]
    (2) edge node [below ] {} (3)
    (2) edge  [bend left,red] node[above ] {$v_2$} (3);

\path[every node/.style={font=\sffamily\Large\bfseries}]
    (3) edge node [below ] {} (4)
    (3) edge  [bend left,red] node[above ] {$v_3$} (4);

\path[every node/.style={font=\sffamily\Large\bfseries},red]
    (4) edge node [below ] {$f$} (5);

\path[every node/.style={font=\sffamily\Large\bfseries},red]
    (5) edge node [below ] {$g$} (10);

\path[every node/.style={font=\sffamily\Large\bfseries}]
    (10) edge node [below ] {} (11)
    (10) edge  [bend left,red] node[above ] {$-v_3$} (11);

\path[every node/.style={font=\sffamily\Large\bfseries}]
    (11) edge node [below ] {} (12)
    (11) edge  [bend left,red] node[above ] {$-v_2$} (12);

 \path[every node/.style={font=\sffamily\Large\bfseries}]
    (12) edge node [below ] {} (13)
    (12) edge  [bend left,red] node[above ] {$-v_1$} (13);

\path[every node/.style={font=\sffamily\Large\bfseries}]
    (6) edge node [right ] {} (2);

\path[every node/.style={font=\sffamily\Large\bfseries}]
    (7) edge node [right ] {} (3);

\path[every node/.style={font=\sffamily\Large\bfseries}]
    (8) edge node [right ] {} (4);

\path[every node/.style={font=\sffamily\Large\bfseries}]
    (9) edge node [right ] {} (5);

\path[every node/.style={font=\sffamily\Large\bfseries}]
    (15) edge node [right ] {} (10);

\path[every node/.style={font=\sffamily\Large\bfseries}]
    (16) edge node [right ] {} (11);

\path[every node/.style={font=\sffamily\Large\bfseries}]
    (17) edge node [right ] {} (12);

\path[every node/.style={font=\sffamily\Large\bfseries}]
    (18) edge node [right ] {} (13);

%\path[every node/.style={font=\sffamily\Large}]
%    (1) edge  [bend right,dashed ] node[above ] {data} (5);

\tikzstyle{every to}=[draw,dashed]
\draw[dashed] (1) to[out=-90,in=-90,looseness=0.1,style={font=\sffamily\Large,dashed}] node[below] {$(X,X)$} (13);

\end{tikzpicture}}
}
Select $\dim(x)=\dim(y)$, $\dim(z)<\dim(x)$ and let the kernels associated with $f$ and $g$ be linear kernels. Then the proposed approach seeks to learn a deformation $\phi_L=(I+v_L)\circ \cdots \circ (I+v_1)$ of the input space such that linear SVD performs well on that input space. Note that
the second portion of the diagram corresponds to the map $(I-v_1)\circ \cdots \circ (I-v_L)$ which is an approximation of the inverse of $\phi_L$ (and an exact inverse in the limit $L\rightarrow \infty$).

\subsection{Mode decomposition}\label{seclkjdekjdhdkjhkj}
\begin{figure}[h]
	\begin{center}
			\includegraphics[width=\textwidth]{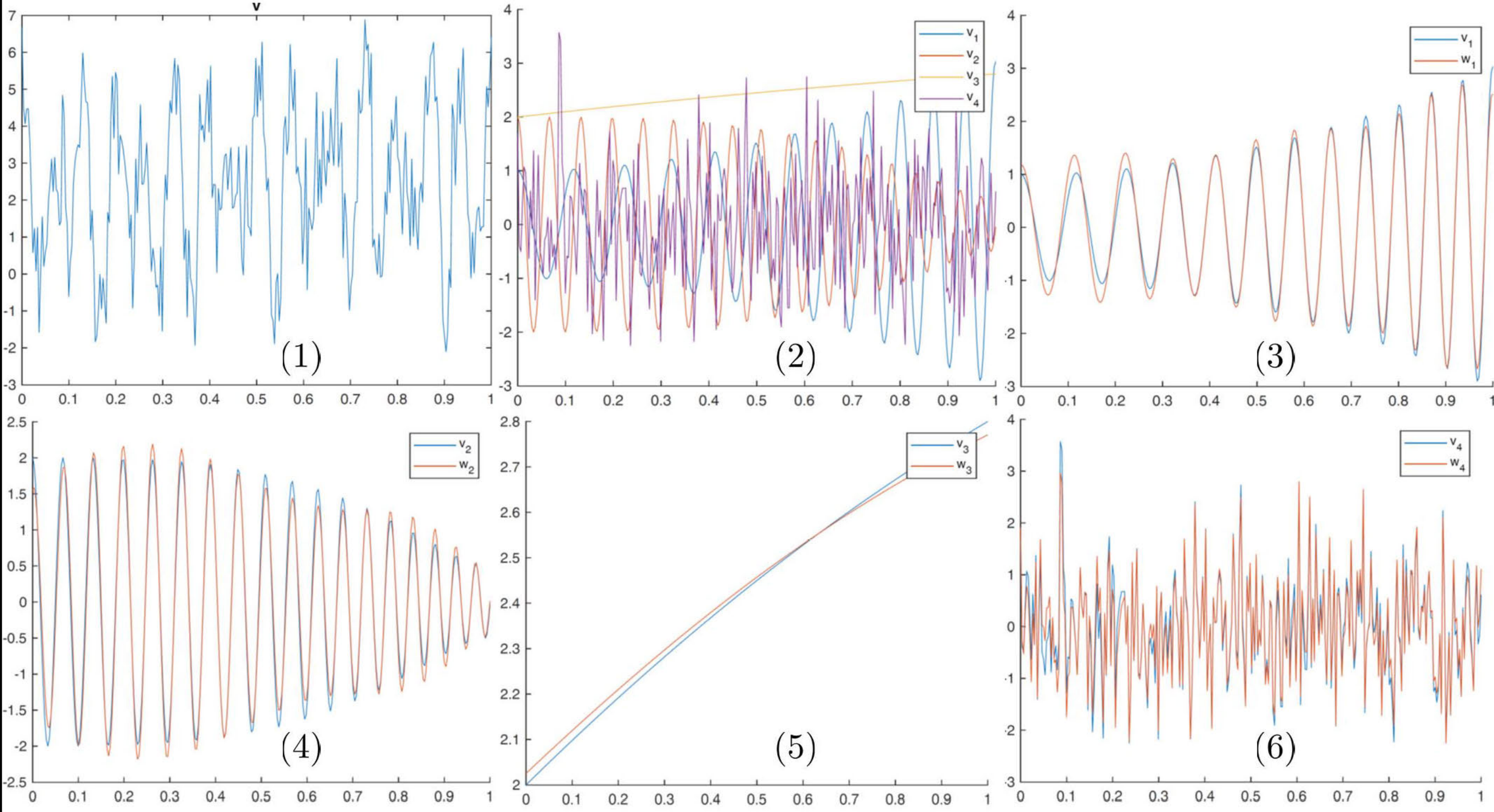}
		\caption{\cite[Fig.~3.2]{owhadi2019kernelmd}. (1) The signal $v=v_1+v_2+v_3+v_4$ (2) The modes $v_1, v_2, v_3, v_4$ (3) $v_1$ and its approximation $w_1$ (4) $v_2$ and its approximation $w_2$ (5) $v_3$ and its approximation $w_3$ (6) $v_4$ and its approximation $w_4$.}\label{figknownfw1}
	\end{center}
\end{figure}

This example is taken from \cite{owhadi2019kernelmd}.
Let $v_1,\ldots,v_m$ be unknown functions mapping $[0,1]$ to $\R$. Consider the problem of recovering $v_1,\ldots,v_m$ given the observation of
$v:=v_1+\cdots+v_m$ at $\bt:=(t_1,\ldots,t_N)\in [0,1]^N$. This mode decomposition problem can naturally be formulated as the CGC problem illustrated in the following diagram for $m=4$.\\
\centerline{
\begin{tikzpicture}[->,>=stealth',shorten >=1pt,auto,node distance=4cm,
                    thick,main node/.style={rectangle,draw,font=\sffamily\Large\bfseries}]

\node[main node] (1) {$t$};
\node[main node] (2) [right of=1] {$v_3(t)$};
\node[main node] (3) [above of=2,node distance=1.5cm] {$v_2(t)$};
\node[main node] (4) [below of=2,node distance=1.5cm] {$v_4(t)$};
\node[main node] (5) [right of=2] {$v(t)$};
\node[main node] (6) [above of=3,node distance=1.5cm] {$v_1(t)$};

\path[every node/.style={font=\sffamily\Large\bfseries},red]
    (1) edge node [above ] {$v_3$} (2);

\path[every node/.style={font=\sffamily\Large\bfseries},red]
    (1) edge node [above ] {$v_2$} (3);

\path[every node/.style={font=\sffamily\Large\bfseries},red]
    (1) edge node [above ] {$v_4$} (4);

\path[every node/.style={font=\sffamily\Large\bfseries},red]
    (1) edge node [above ] {$v_1$} (6);

\path[every node/.style={font=\sffamily\Large\bfseries}]
    (2) edge node [above ] {} (5);

\path[every node/.style={font=\sffamily\Large\bfseries}]
    (3) edge node [above ] {} (5);

\path[every node/.style={font=\sffamily\Large\bfseries}]
    (4) edge node [above ] {} (5);

\path[every node/.style={font=\sffamily\Large\bfseries}]
    (6) edge node [above ] {} (5);

\tikzstyle{every to}=[draw,dashed]
\draw[dashed] (1) to[out=-90,in=-90,looseness=1,style={font=\sffamily\Large\bfseries,dashed}] node[above] {$(\bt, v(\bt))$} (5);

\end{tikzpicture}}
Given kernels $K_1, \ldots, K_m$, the GP solution is then to recover $v_1,\ldots,v_m$ with the minimizer of
\begin{equation}\label{eqhgeerewwt2yv}
\begin{cases}
\minimize_{v_1,\ldots,v_m}&\sum_{i=1}^m\|v_i\|_{K_i}^2\\
\st&\sum_{i=1}^m v_i(\bt)=v(\bt)\,.
\end{cases}
\end{equation}
Writing $K:=\sum_{i=1}^m K_i$ and $w_i$ for the minimizer over $v_i$ of
 \eqref{eqhgeerewwt2yv}, we have
 \begin{equation}
 w_i(t)=K_i(t,\bt) K(\bt,\bt)^{-1}v(\bt)\,.
 \end{equation}
Evidently, the selection of the kernels $K_i$ depends on prior information on the modes $v_i$.
For instance, consider the case where, $m=4$,  $v_1(t)=a_1(t)\cos(\theta_1(t))+b_1(t) \sin(\theta_1(t)) $, $v_2(t)=a_2(t)\cos(\theta_2(t))+b_2(t)\sin(\theta_2(t)) $, $v_3(t)=a_3(t)$,   $v_4$ is white-noise, the $a_i$ and $b_i$ are unknown smooth functions and the $\theta_i$ are known.
 Replacing $a_1,b_1,a_2,b_2$ by independent centered GPs with Gaussian kernel $e^{-\frac{(s-t)^2}{\gamma^2}}$ results in the  selection
\begin{equation}
K_1(s,t)= e^{-\frac{(s-t)^2}{\gamma^2}}\big(\cos(\theta_1(s))\cos(\theta_1(t))+
\sin(\theta_1(s))\sin(\theta_1(t))\big)
\end{equation}
 and
\begin{equation}
K_2(s,t)= e^{-\frac{(s-t)^2}{\gamma^2}}\big(\cos(\theta_2(s))\cos(\theta_2(t))+\sin(\theta_2(s))
\sin(\theta_2(t))\big).
\end{equation}
 For $v_3$ we select
 $K_3(s,t)= 1+st+e^{-\frac{(s-t)^2}{4}}$. Finally since $v_4$ is white noise we use
$K_4(s,t)= \updelta(s-t)$.
 Fig.~\ref{figknownfw1} illustrates the proposed approach ($\bt$ corresponds to $320$ points distributed over a grid of $[0,1]$).
Note that the selection of $K_1$ and $K_2$ can also be formalized by unpacking $v_1$ and $v_2$  into subgraphs (illustrated below for $v_1$) involving the unknown functions $a_i, b_i$ and the known functions $\cos(\theta_i), \sin(\theta_i)$.\\
\centerline{
\begin{tikzpicture}[->,>=stealth',shorten >=1pt,auto,node distance=3cm,
                    thick,main node/.style={rectangle,draw,font=\sffamily\Large\bfseries}]

\node[main node] (1) {$t$};
\node[main node] (2) [right of=1] {$\theta_1(t)$};
\node[main node] (3) [above of=2,node distance=2cm] {$a_1(t)$};
\node[main node] (4) [below of=2,node distance=2cm] {$b_1(t)$};
\node[main node] (5) [right of=3,circle] {};
\node[main node] (6) [right of=4,circle] {};
\node[main node] (7) [below of=5,node distance=1.5cm] {$\cos \theta_1$};
\node[main node] (8) [above of=6,node distance=1.5cm] {$\sin \theta_1$};
\node[main node] (9) [right of=5] {$a_1\cos \theta_1$};
\node[main node] (10) [right of=6] {$b_1\sin \theta_1$};
\node[main node] (11) [below of=9,node distance=2cm] {$v_1$};

%\node[main node] (7) [right of=2] {};

\path[every node/.style={font=\sffamily\Large\bfseries}]
    (1) edge node [above ] {$\theta_1$} (2);

\path[every node/.style={font=\sffamily\Large\bfseries},red]
    (1) edge node [above ] {$a_1$} (3);

\path[every node/.style={font=\sffamily\Large\bfseries},red]
    (1) edge node [above ] {$b_1$} (4);

\path[every node/.style={font=\sffamily\Large\bfseries}]
    (2) edge node [above ] {$\cos$} (7);

\path[every node/.style={font=\sffamily\Large\bfseries}]
    (2) edge node [below ] {$\sin$} (8);

\path[every node/.style={font=\sffamily\Large\bfseries}]
    (3) edge node [above ] {} (5);

\path[every node/.style={font=\sffamily\Large\bfseries}]
    (4) edge node [above ] {} (6);

\path[every node/.style={font=\sffamily\Large\bfseries}]
    (7) edge node [above ] {} (5);

\path[every node/.style={font=\sffamily\Large\bfseries}]
    (8) edge node [above ] {} (6);

\path[every node/.style={font=\sffamily\Large\bfseries}]
    (5) edge node [above ] {} (9);

\path[every node/.style={font=\sffamily\Large\bfseries}]
    (6) edge node [above ] {} (10);

\path[every node/.style={font=\sffamily\Large\bfseries}]
    (9) edge node [above ] {} (11);

\path[every node/.style={font=\sffamily\Large\bfseries}]
    (10) edge node [above ] {} (11);

\end{tikzpicture}}

   \begin{figure}[h]
	\begin{center}
			\includegraphics[width=\textwidth]{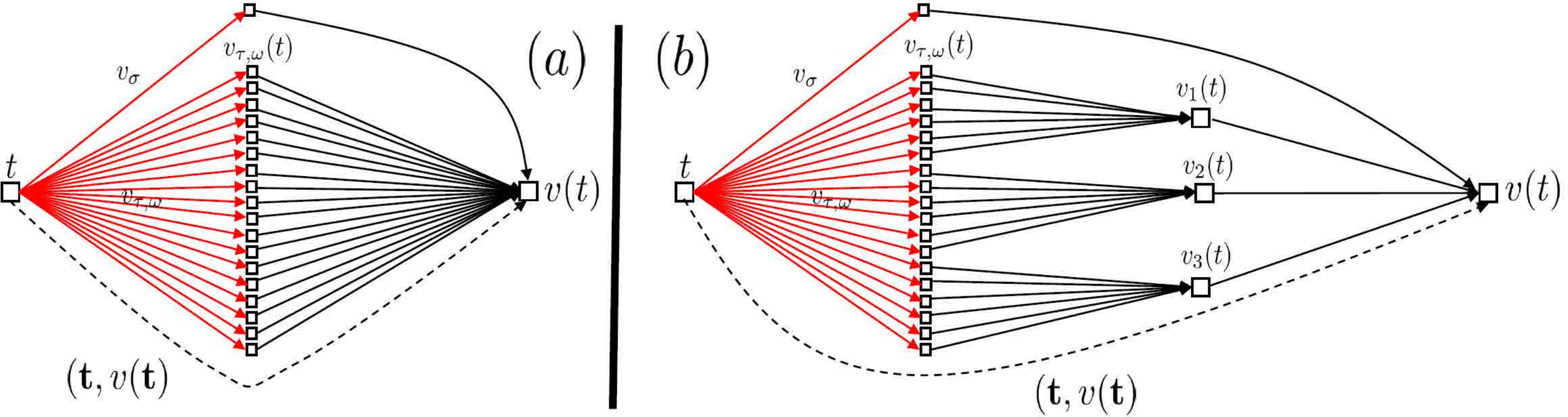}
		\caption{CGC diagrams for empirical mode decomposition. (a) Recovery of the fine modes $v_{\tau,\omega}$ (b) Recovery of the coarse modes $v_i$ through aggregation of the fine modes $v_i$.}\label{figknownfw1b}
	\end{center}
\end{figure}
  \begin{figure}[h]
        \begin{center}
                        \includegraphics[width=\textwidth]{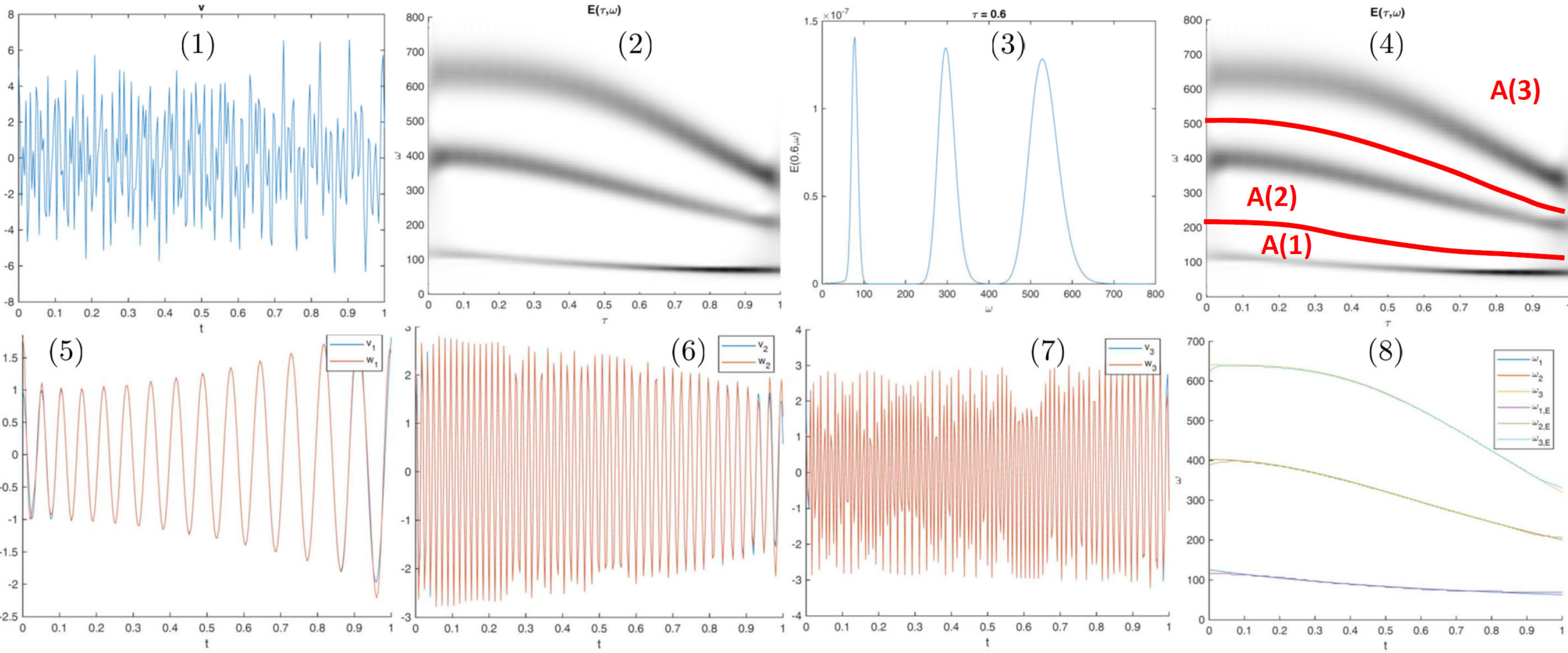}
                \caption{\cite[Fig.~4.10]{owhadi2019kernelmd}. (1) The signal $v=u+v_\sigma$ where $u=v_1+v_2+v_3$, $v_\sigma\sim \cN(0, \sigma^2 \updelta(t-s))$  and $\sigma=0.01$ (2) $(\tau,\omega)\rightarrow E(\tau,\omega)$ defined by \eqref{eqkndddenkjednd} (one can identify three stripes) (3) $\omega\rightarrow E(0.6,\omega)$ (4)
Partitioning $[0,1]\times [\omega_{\min},\omega_{\max}]=\cup_{i=1}^3 A_i$ of the time frequency domain into three disjoint subsets identified from $E$ (5) $v_1$ and its approximation $w_1$ (6) $v_2$ and its approximation $w_2$ (7)
$v_3$ and its approximation $w_3$ (8) $\omega_1,\omega_2,\omega_3$ and their approximations $\omega_{1,E},\omega_{2,E},\omega_{3,E}$.}\label{figunknownf2}
        \end{center}
\end{figure}

\subsection{Dynamic Computational Graph Completion}\label{Subsectd}

The process, presented in Sec.~\ref{seclkjdekjdhdkjhkj}, of programming kernels via a computational graph by decomposing and recomposing simpler kernels is generic  \cite{owhadi2019kernelmd} and can be employed to design the computation graph itself.
As an example, using the notations of Sec.~\ref{seclkjdekjdhdkjhkj}, consider the classical empirical mode decomposition problem \cite{huang1998empirical} of recovering $(v_1,\ldots,v_m)$ (Fig.~\ref{figunknownf2}.5-7) from the data $(\bt, v(\bt))$ (Fig.~\ref{figunknownf2}.1) where $v=v_\sigma+\sum_{i=1}^m v_i$.
In that problem,  $m$ is unknown, $v_\sigma$ is white noise and the $v_i$ are of the form $v(t)=a_i(t)\cos\big(\theta_i(t)\big)$ where the
 $a_i$ are piecewise smooth unknown functions and the instantaneous frequencies $\dot{\theta}_i$ are   strictly positive, well separated and unknown smooth functions.

Then, as shown in \cite{owhadi2019kernelmd},  the $v_i$ can be recovered through a hierarchical and dynamic programming approach to the design of a computational  graph (of the type presented in Sec.~\ref{seclkjdekjdhdkjhkj}). In this approach the $v_i$ are identified as aggregates (integrals/sums, see Fig.~\ref{figknownfw1b}.(b))
 of finer/simpler modes, i.e.
 \begin{equation}
 v_i(t)=\sum_{(\tau,\omega)\in A_i} v_{\tau,\omega}(t)\,,
 \end{equation}
where the $A_i$ are an unknown partition of the time-frequency domain spanned by $(\tau,\omega)$, and,
  \begin{equation}
v_{\tau,\omega}(t)=a_{\tau,\omega} \chi_{\tau,\omega,c}(t)+b_{\tau,\omega} \chi_{\tau,\omega,s}(t)
 \end{equation}
where the $a$ and $b$ are unknown functions and the $\chi$ are Gabor wavelets defined by
\begin{eqnarray}\label{eqjhgjguyggt}
\chi_{\tau,\omega,c}(t)&:=&\Bigl(\frac{2}{\pi}\Bigr)^\frac{1}{4} \sqrt{\frac{\omega}{\alpha}}\cos(\omega (t-\tau))e^{-\frac{ \omega^2 (t-\tau)^2}{\alpha^2}}\,, \quad t \in \R, \nonumber\\
\chi_{\tau,\omega,s}(t)&:=&\Bigl(\frac{2}{\pi}\Bigr)^\frac{1}{4} \sqrt{\frac{\omega}{\alpha}}\sin(\omega (t-\tau))e^{-\frac{ \omega^2 (t-\tau)^2}{\alpha^2}}\,, \quad t \in \R\,.
\end{eqnarray}
Observing that
\begin{equation}
\sum_{(\tau,\omega)}  v_{\tau,\omega}(t)=v(t)\,,
\end{equation}
the recovery of the unknown functions $a$ and $b$ is a linear mode decomposition problem whose CGC diagram is illustrated in Fig.~\ref{figknownfw1b}.(a).
Replacing $a$ and $b$ by white noise leads to the approximation of $v_{\tau,\omega}$ with
\begin{equation}
w_{\tau,\omega}(t)=\sum_{\tau,\omega} K_{\tau,\omega}(t,\bt) K(\bt,\bt)^{-1} v(\bt)\,,
\end{equation}
where $K=\sum_{\tau,\omega} K_{\tau,\omega}$ and
\begin{equation}
K_{\tau,\omega}(t,t')=\chi_{\tau,\omega,c}(t) \chi_{\tau,\omega,c}(t')+ \chi_{\tau,\omega,s}(t) \chi_{\tau,\omega,s}(t')\,.
\end{equation}
To identify the unknown partition formed by the subsets $A_i$ (i.e., the connectivity of the graph illustrated in Fig.~\ref{figknownfw1b}.(b)), we compute the inner product
\begin{equation}\label{eqkndddenkjednd}
E(\tau,\omega)=    (w_{\tau,\omega}(\bt))^T K(\bt,\bt)^{-1}v(\bt)\,.
\end{equation}
This inner product  is the variance of $(v_{\tau,\omega}(\bt))^T K(\bt,\bt)^{-1}v(\bt)$ when  the coefficients $a_{\tau,\omega}$ and $b_{\tau,\omega}$ of $v_{\tau,\omega}$ are randomized as white noise and  can be interpreted as the level of activation of the resulting  GP $v_{\tau,\omega}$ after conditioning on the data.

Figure \ref{figunknownf2} (\cite[Fig.~4.10]{owhadi2019kernelmd}) illustrates the proposed approach for a noisy signal composed of three quasi-trigonometric modes
 (the fourth mode is white noise).
Figure \ref{figunknownf2}.1  displays the total observed signal $v$ and the three quasi-trigonometric modes
$v_{1},v_{2}, v_{3} $ constituting $v$
 are displayed in Figures \ref{figunknownf2}.5, 6  and 7,  along with their recoveries
$w_{1},w_{2}$  and $w_{3}$.
 Figure \ref{figunknownf2}.8 also shows  approximations of the instantaneous frequencies obtained as
\begin{equation}\label{eqkjhjhedbdjhdb}
\omega_{i,E}(t):=\operatorname{argmax}_{\omega: (t,\omega)\in A_i} E(t,\omega)\,.
\end{equation}

\begin{Remark}
{
The approach \eqref{eqkndddenkjednd} of computing the level of activation of GP in the graph is generic (it can be generalized to other graphs with multiple paths between two nodes) and can be used to induce a ranking and a pruning of the edges (unknown functions) of the graph \cite{owhadi2019kernelmd}.
}
\end{Remark}

\subsection{Cyclic Graph}\label{seccyc}
In the following example, taken from \cite{owhadi2019kernelmd}, we illustrate a CGC problem involving a cyclic graph. The underlying algorithm is used in  \cite[Chap.~6]{owhadi2019kernelmd} to solve empirical mode decomposition problems (as described in Sec.~\ref{Subsectd}, with possibly unknown non-trigonometric waveforms)  to near machine precision.
The CGC diagram of this algorithm is illustrated by the following diagram.

\begin{equation}\label{eqkjwdleddhdguye}
\begin{tikzpicture}[->,>=stealth',shorten >=1pt,auto,node distance=2cm,
                    thick,main node/.style={rectangle,draw,font=\sffamily\Large\bfseries}]

\node[main node] (1) {$t$};
\node[main node] (2) [right of=1] {$\theta_e(t)$};
\node[main node] (3) [right of=2] {$f(t)$};
\node[main node] (4) [above of=3,node distance=1.5cm] {$\chi_c(t)$};
\node[main node] (5) [below of=3,node distance=1.5cm] {$\chi_s(t)$};
\node[main node] (6) [right of=3,blue] {$w$};
\node[main node] (7) [above of=6,node distance=1.5cm,circle] {};
\node[main node] (8) [below of=6,node distance=1.5cm,circle] {};
\node[main node] (9) [right of=7,blue] {$Z_c$};
\node[main node] (10) [right of=8,blue] {$Z_s$};
\node[main node] (11) [right of=6,circle] {};
\node[main node] (12) [right of=11] {$\delta \theta_e(t)$};

\path[every node/.style={font=\sffamily\Large\bfseries},red]
    (1) edge node [above ] {$\theta_e$} (2);

\path[every node/.style={font=\sffamily\Large}]
    (1) edge  [bend right,dashed ] node[below ] {$f$} (3);

\path[every node/.style={font=\sffamily\Large\bfseries}]
    (2) edge node [above ] {} (4)
    (2) edge node [above ] {} (5)
    (4) edge node [above ] {} (7)
    (5) edge node [above ] {} (8)
    (7) edge node [above ] {} (3)
    (8) edge node [above ] {} (3);

\path[every node/.style={font=\sffamily\Large\bfseries},blue]
    (6) edge node [above] {} (3);

\path[every node/.style={font=\sffamily\Large\bfseries},blue]
    (9) edge node [above] {} (7)
    (10) edge node [above] {} (8);

\path[every node/.style={font=\sffamily\Large\bfseries}]
    (9) edge node [above] {} (11)
    (10) edge node [above] {} (11)
    (11) edge node [above] {} (12);

\tikzstyle{every to}=[draw,dashed]
\draw (12) to[out=90,in=90,looseness=1,style={font=\sffamily\Large\bfseries}] node[above] {} (2);

\end{tikzpicture}
\end{equation}

To describe it, let $\tau$ a given time, $\theta_{e}$ be an estimated phase function (to be refined), and a signal $f$. Suppose that $f$ contains a mode of the form $ a(t) \cos(\theta(t))$  near the time $\tau$. The problem is to estimate $a$ and $\theta$ near $\tau$.
The cycle in the graph allows us to do so by successively correcting/refining an initial estimate $\theta_e$ of $\theta$.
This is done by identifying an updated estimate $ \theta_{e}(\tau) +\delta \theta (\tau)$ of $\theta$ by solving the following GP regression problem
\begin{equation}
\label{def_vwindow}
\big(f(t) - X_c \cos(\theta_e(t)-X_s\sin(\theta_e(t))-w(t)\big) e^{-\big(\frac{\dot{\theta_{e}}(\tau)(t - \tau)}{\alpha}\big)^2} =0\,,
\end{equation}
where $w$ is white noise and $X_c$ and $X_s$ are independent $\cN(0,1)$ random variables.
$a$ and the correction $\delta \theta$ are then identified as

\begin{eqnarray}
\label{eienuriig}
a(\tau)&:=&\sqrt{X^{2}_{c}+X^2_s}\,\nonumber\\
 \delta\theta(\tau)&: = &\operatorname{atan2}\big(-X_{s}, X_c)\big)\,.
\end{eqnarray}
\begin{Remark}
{
Although some computational graphs may be oriented as in Sec.~\ref{subsecekjhdedgh} or Deep Learning with information flowing from one side to the other, others as in \eqref{eqkjwdleddhdguye} may be cyclic (and in particular not oriented).
}
\end{Remark}

\section{Concluding remarks}

{ We will now point towards interesting development directions for CGC.   }

\subsection{Incorporating symmetries}
Unknown functions can be informed about underlying symmetries by employing equivariant kernels  \cite{reisert2007learning} and their generalization \cite[Sec.~9.2]{owhadi2020ideas}. Such kernels are can be obtained by averaging a given kernel with respect to the action of a unitary group of transformations. From the perspective of CGC, convolutional neural networks can be recovered (see \cite[Sec.~9.2]{owhadi2020ideas}) by replacing unknown functions with GPs of the form
\begin{equation}\label{eqkjhehd}
\xi=\frac{1}{|\G|}\sum_{g\in \G} g^T R^T \xi_g(P g x)
\end{equation}
where the $\xi_g$ are independent GPs, $\G$ is a finite group of unitary transformations, $|\G|$ is its volume with respect to the Haar measure, $P$ and $R$ are linear projection operators.

\subsection{Error estimates}
{
Using kernels places CGC on solid mathematical foundations and opens the possibility of computing error bounds on the recovery of unknown functions in terms of their assumed regularity expressed as a bound on their RKHS norm.
In the interpolation setting of Sec.~\ref{secinterpset}, these error estimates take the form \cite{Wendland:2005}
\begin{equation}
\big|f(x)-f^\dagger(x)\big|\leq \sigma(x) \|f\|_K\,
\end{equation}
where $f$ is the true function, $f^\dagger$ is its approximation \eqref{eqhgvygvgyv2}, and $\sigma(x)$ is the conditional standard deviation of the GP $\xi \sim \cN(0,K)$ given $\xi(X)=Y$, which can be identified via $\sigma^2(x)=K(x,x)-K(x,X)K(X,X)^{-1}K(X,x)$.
In the generalized interpolation setting of Sec.!\ref{seckejddjdhkj}, these error estimates take the form \cite[Thm.~5.1]{Owhadi:2014}
\begin{equation}
\big|[\varphi,u]-[\varphi,u^\dagger]\big|\leq \sigma(\varphi) \|f\|_K\,
\end{equation}
with  $\sigma^2(\varphi)=K(\varphi,\varphi)-K(\varphi,\phi)K(\phi,\phi)^{-1}K(\phi,\varphi)$.
In the idea registration setting of Sec.~\ref{subsecekjhdedgh}, if the data $(X_i,Y_i)$ is produced by a true function $F$ ($Y_i=F(X_i)$), then these error estimates take the form  \cite[Cor.~8.10]{owhadi2020ideas}
\begin{equation}
\big|F(x)-f\circ \phi_L(x)\big|\leq \sigma^v(x) \|F\|_{K^v}\,
\end{equation}
where $\phi_L=(I+v_L)\circ \cdots \circ (I+v_1)$, $(f,v_1,\ldots,v_L)$ is a minimizer of \eqref{eqhjejhdejhddh}, $K^{v}$ is the warped kernel
$ K^{v}(x,x')=K(\phi_L(x),\phi_L(x'))$ and $\sigma^v(x)=K^v(x,x)-K^v(x,X)K^v(X,X)^{-1}K^v(X,x)$.
In the setting of PDEs, more refined estimated can be obtained by exploiting Poincar\'{e} inequalities and the continuity of the underlying differential operators \cite{OwhScobook2018}, \cite[Prop.~5.2]{chen2021solving}.
Developing general error estimates for CGC (e.g., for minimizers of \eqref{eqhdewdgvgyv}) is beyond the scope of this paper.

}

\subsection{Learning kernels for CGC}
{

We have focused the manuscript on the situation where the kernels of the underlying GPs are given/pre-determined.
Using data-driven kernels can improve the accuracy of kernel methods by orders of magnitude \cite{owhadi2019kernel, chen2021consistency, hamzi2021learning, hamzi2021simple, darcy2021learning, lee2021learning, prasanth2021kernel}.
There are essentially three categories of methods for learning kernel from data: variants of cross-validation (such as Kernel Flows \cite{owhadi2019kernel}), maximum likelihood estimation (see \cite{chen2021consistency} for a comparison between Kernel Flows and MLE), and maximum a posteriori estimation \cite{owhadi2020ideas}.
We refer to \cite{darcy2022learningcgc} for the description of a randomized cross-validation approach for learning kernels in CGC (in the setting of one-shot/trajectory learning SDEs from data, in that setting the some of the noise variables of the Computational Graph represent the  Brownian motion increments driving the SDE).
}

\subsection{Computational hypergraph completion and constraints}
{

The CGC framework is equivalent to the completion of oriented hypergraphs representing functional dependencies between (input/output) groups of variables.
Recall \cite{Rusnak2013} that an oriented hypergraph is a tripe $G=(\V,\cE,\varphi)$ such that $\V$ is a set of vertices, $\cE$ is a multiset of edges  (a set of subsets of $\V$), $\varphi \,:\, \V \times \cE\rightarrow \{-1,0,1\}$ (known as an incidence function) is such that
$\varphi(i,e)\not=0 \Leftrightarrow i\in e$, $\varphi(i,e)=1$ if vertex $i$ is an input for the edge $e$ and $\varphi(i,e)=-1$ if vertex $i$ is an output for the edge $e$. Let each vertex $i\in \V$  represent a variable $v_i$ and  let each edge $e\in \cE$ represent a function $f_e$. Then the oriented hypergraph can be used to represent  the following functional identities between groups of input variables and output variables.
\begin{equation}
f_e\big( \otimes_{\{\varphi(i,e)=1\}} v_i)=\otimes_{\{\varphi(j,e)=-1\}} v_j \text{ for } e\in \cE\,,
\end{equation}
where $\otimes_{\{\varphi(i,e)=1\}} v_i$ is the tensorization of the variables $v_i$ such that $\varphi(i,e)=1$.
Then unknown functions can, as  before, be recovered by randomizing them through Gaussian priors and computing their MAP estimator.
Therefore, by employing an oriented hypergraph representation of a CSE problem, one can restrict the elements of $\V$ to represent primary variables and not introduce round nodes representing tensorized/aggregated variables. Here we have employed the graph representation (with aggregated variables) to facilitate visual representation.
By employing operator-valued kernels, the proposed framework can be generalized to the situation where the elements of $\V$ and $\cE$ are functions and functionals (nonlinear operators acting on spaces of functions).
The CGC framework can also be extended to (possibly non-oriented) hypergraphs representing implicit functional dependencies between groups of variables. Given a hypergraph $G=(\V,\cE)$ ($\V$ is a set of vertices and $\cE$ is a set of subsets of $\V$), each vertex $i\in \V$  represents a variable $v_i$ and each edge $e\in \cE$ represents a function $f_e$ and the  hypergraph can be used to represent the following functional implicit equality or inequality constraints between variables.
\begin{equation}
f_e\big( \otimes_{i\in e} v_i)=0 \text{ for } e\in \cE_1 \text{ and } f_e\big( \otimes_{i\in e} v_i)\geq 0 \text{ for } e\in \cE_2\,.
\end{equation}
Then unknown functions can, as before, be recovered by randomizing them through Gaussian priors and computing their MAP estimator.
}

\subsection*{Data availability statement}
Data sharing is not applicable to this article as no new data were created or analyzed in this study.

\subsection*{Acknowledgments}
The author gratefully acknowledges partial support by the Air Force Office of Scientific Research under MURI award number FA9550-20-1-0358 (Machine Learning and Physics-Based Modeling and Simulation). Thanks to Amy Braverman, Jouni Susiluoto, and  Otto Lamminpaeae for stimulating discussions.
Thanks to an anonymous referee and to Jean-Luc Cambier for helpful comments and feedback.

\bibliographystyle{plain}
\bibliography{merged,RPS,extra,kmd}

\def\cprime{$'$} \def\cprime{$'$} \def\cprime{$'$} \def\cprime{$'$}
  \def\cprime{$'$}
\begin{thebibliography}{10}

\bibitem{baldi2012autoencoders}
Pierre Baldi.
\newblock Autoencoders, unsupervised learning, and deep architectures.
\newblock In {\em Proceedings of ICML workshop on unsupervised and transfer
  learning}, pages 37--49. JMLR Workshop and Conference Proceedings, 2012.

\bibitem{belkin2021fit}
Mikhail Belkin.
\newblock Fit without fear: remarkable mathematical phenomena of deep learning
  through the prism of interpolation.
\newblock {\em arXiv preprint arXiv:2105.14368}, 2021.

\bibitem{brown1992survey}
Lisa~Gottesfeld Brown.
\newblock A survey of image registration techniques.
\newblock {\em ACM computing surveys (CSUR)}, 24(4):325--376, 1992.

\bibitem{chen2021solving}
Yifan Chen, Bamdad Hosseini, Houman Owhadi, and Andrew~M Stuart.
\newblock Solving and learning nonlinear pdes with gaussian processes.
\newblock {\em Journal of Computational Physics}, 2021.
\newblock arXiv preprint arXiv:2103.12959.

\bibitem{chen2021consistency}
Yifan Chen, Houman Owhadi, and Andrew Stuart.
\newblock Consistency of empirical bayes and kernel flow for hierarchical
  parameter estimation.
\newblock {\em Mathematics of Computation}, 2021.

\bibitem{cockayne2019bayesian}
Jon Cockayne, Chris~J Oates, Timothy~John Sullivan, and Mark Girolami.
\newblock Bayesian probabilistic numerical methods.
\newblock {\em SIAM Review}, 61(4):756--789, 2019.

\bibitem{constantine2014active}
P.~G. Constantine, E.~Dow, and Q.~Wang.
\newblock Active subspace methods in theory and practice: applications to
  kriging surfaces.
\newblock {\em SIAM Journal on Scientific Computing}, 36(4):A1500--A1524, 2014.

\bibitem{cressie1988spatial}
Noel Cressie.
\newblock Spatial prediction and ordinary kriging.
\newblock {\em Mathematical geology}, 20(4):405--421, 1988.

\bibitem{darcy2022learningcgc}
Matthieu Darcy, Boumediene Hamzi, Giulia Livieri, Houman Owhadi, and Peyman
  Tavallali.
\newblock One-shot learning of stochastic differential equations with
  computational graph completion, 2022.

\bibitem{darcy2021learning}
Matthieu Darcy, Boumediene Hamzi, Jouni Susiluoto, Amy Braverman, and Houman
  Owhadi.
\newblock Learning dynamical systems from data: a simple cross-validation
  perspective, part ii: nonparametric kernel flows, 2021.

\bibitem{dunlop2020hyperparameter}
Matthew~M Dunlop, Tapio Helin, and Andrew~M Stuart.
\newblock Hyperparameter estimation in bayesian map estimation:
  parameterizations and consistency.
\newblock {\em The SMAI journal of computational mathematics}, 6:69--100, 2020.

\bibitem{fensel2020knowledge}
Dieter Fensel, U~Simsek, Kevin Angele, Elwin Huaman, Elias K{\"a}rle,
  Oleksandra Panasiuk, Ioan Toma, J{\"u}rgen Umbrich, and Alexander Wahler.
\newblock {\em Knowledge graphs}.
\newblock Springer, 2020.

\bibitem{golub1971singular}
Gene~H Golub and Christian Reinsch.
\newblock Singular value decomposition and least squares solutions.
\newblock In {\em Linear algebra}, pages 134--151. Springer, 1971.

\bibitem{grenander1998computational}
Ulf Grenander and Michael~I Miller.
\newblock Computational anatomy: An emerging discipline.
\newblock {\em Quarterly of applied mathematics}, 56(4):617--694, 1998.

\bibitem{hamzi2021simple}
B~Hamzi, R~Maulik, and H~Owhadi.
\newblock Simple, low-cost and accurate data-driven geophysical forecasting
  with learned kernels.
\newblock {\em Proceedings of the Royal Society A}, 477(2252):20210326, 2021.

\bibitem{hamzi2021learning}
Boumediene Hamzi and Houman Owhadi.
\newblock Learning dynamical systems from data: A simple cross-validation
  perspective, part i: Parametric kernel flows.
\newblock {\em Physica D: Nonlinear Phenomena}, 421:132817, 2021.

\bibitem{he2016deep}
Kaiming He, Xiangyu Zhang, Shaoqing Ren, and Jian Sun.
\newblock Deep residual learning for image recognition.
\newblock In {\em Proceedings of the IEEE conference on computer vision and
  pattern recognition}, pages 770--778, 2016.

\bibitem{Hennig2015}
P.~Hennig, M.~A. Osborne, and M.~Girolami.
\newblock Probabilistic numerics and uncertainty in computations.
\newblock {\em Proc. R. Soc. A.}, 471(2179):20150142, 2015.

\bibitem{huang1998empirical}
N.~E. Huang, Z.~Shen, S.~R. Long, M.~C. Wu, H.~H. Shih, Q.~Zheng, N.-C. Yen,
  C.~C. Tung, and H.~H. Liu.
\newblock The empirical mode decomposition and the {H}ilbert spectrum for
  nonlinear and non-stationary time series analysis.
\newblock {\em Proceedings of the Royal Society of London. Series A:
  Mathematical, Physical and Engineering Sciences}, 454(1971):903--995, 1998.

\bibitem{jordan1998learning}
Michael~Irwin Jordan.
\newblock {\em Learning in graphical models}, volume~89.
\newblock Springer Science \& Business Media, 1998.

\bibitem{lee2021learning}
Jonghyeon Lee, Edward De~Brouwer, Boumediene Hamzi, and Houman Owhadi.
\newblock Learning dynamical systems from data: A simple cross-validation
  perspective, part iii: Irregularly-sampled time series.
\newblock {\em arXiv preprint arXiv:2111.13037}, 2021.

\bibitem{lin2015learning}
Yankai Lin, Zhiyuan Liu, Maosong Sun, Yang Liu, and Xuan Zhu.
\newblock Learning entity and relation embeddings for knowledge graph
  completion.
\newblock In {\em Twenty-ninth AAAI conference on artificial intelligence},
  2015.

\bibitem{micchelli1977survey}
C.~A. Micchelli and T.~J. Rivlin.
\newblock A survey of optimal recovery.
\newblock In {\em Optimal Estimation in Approximation Theory}, pages 1--54.
  Springer, 1977.

\bibitem{mika1998kernel}
Sebastian Mika, Bernhard Sch{\"o}lkopf, Alexander~J Smola, Klaus-Robert
  M{\"u}ller, Matthias Scholz, and Gunnar R{\"a}tsch.
\newblock Kernel pca and de-noising in feature spaces.
\newblock In {\em NIPS}, volume~11, pages 536--542, 1998.

\bibitem{noy2019industry}
Natasha Noy, Yuqing Gao, Anshu Jain, Anant Narayanan, Alan Patterson, and Jamie
  Taylor.
\newblock Industry-scale knowledge graphs: lessons and challenges.
\newblock {\em Communications of the ACM}, 62(8):36--43, 2019.

\bibitem{OwhScobook2018}
H.~Owhadi and C.~Scovel.
\newblock {\em Operator Adapted Wavelets, Fast Solvers, and Numerical
  Homogenization, from a game theoretic approach to numerical approximation and
  algorithm design}.
\newblock Cambridge Monographs on Applied and Computational Mathematics.
  Cambridge University Press, 2019.

\bibitem{OwhScoSchNotAMS2019}
H.~Owhadi, C.~Scovel, and F.~Sch\"{a}fer.
\newblock Statistical {N}umerical {A}pproximation.
\newblock 66(10), 2019.

\bibitem{Owhadi:2014}
Houman Owhadi.
\newblock Bayesian numerical homogenization.
\newblock {\em Multiscale Modeling \& Simulation}, 13(3):812--828, 2015.

\bibitem{owhadi2017multigrid}
Houman Owhadi.
\newblock Multigrid with rough coefficients and multiresolution operator
  decomposition from hierarchical information games.
\newblock {\em SIAM Review}, 59(1):99--149, 2017.

\bibitem{owhadi2020ideas}
Houman Owhadi.
\newblock Do ideas have shape? plato's theory of forms as the continuous limit
  of artificial neural networks.
\newblock {\em arXiv preprint arXiv:2008.03920}, 2020.

\bibitem{owhadinotesopvkgp2021}
Houman Owhadi.
\newblock Notes on operator valued kernels, feature maps and gaussian
  processes.
\newblock 2021.
\newblock
  \url{http://users.cms.caltech.edu/~owhadi/index_htm_files/OperatorValuedGPs.pdf}.

\bibitem{owhadi2019operator}
Houman Owhadi and Clint Scovel.
\newblock {\em Operator-Adapted Wavelets, Fast Solvers, and Numerical
  Homogenization: From a Game Theoretic Approach to Numerical Approximation and
  Algorithm Design}, volume~35.
\newblock Cambridge University Press, 2019.

\bibitem{owhadi2019kernelmd}
Houman Owhadi, Clint Scovel, and Gene~Ryan Yoo.
\newblock {\em Kernel Mode Decomposition and the programming of kernels}.
\newblock Springer, 2021.
\newblock arXiv preprint arXiv:1907.08592 for early version.

\bibitem{owhadi2019kernel}
Houman Owhadi and Gene~Ryan Yoo.
\newblock Kernel flows: From learning kernels from data into the abyss.
\newblock {\em Journal of Computational Physics}, 389:22--47, 2019.

\bibitem{prasanth2021kernel}
Sai Prasanth, Ziad~S Haddad, Jouni Susiluoto, Amy~J Braverman, Houman Owhadi,
  Boumediene Hamzi, Svetla~M Hristova-Veleva, and Joseph Turk.
\newblock Kernel flows to infer the structure of convective storms from
  satellite passive microwave observations.
\newblock In {\em AGU Fall Meeting 2021}. AGU, 2021.

\bibitem{raissi2017inferring}
M.~Raissi, P.~Perdikaris, and G.~E. Karniadakis.
\newblock Inferring solutions of differential equations using noisy
  multi-fidelity data.
\newblock {\em Journal of Computational Physics}, 335:736--746, 2017.

\bibitem{raissi2019physics}
Maziar Raissi, Paris Perdikaris, and George~E Karniadakis.
\newblock Physics-informed neural networks: A deep learning framework for
  solving forward and inverse problems involving nonlinear partial differential
  equations.
\newblock {\em Journal of Computational Physics}, 378:686--707, 2019.

\bibitem{reisert2007learning}
Marco Reisert and Hans Burkhardt.
\newblock Learning equivariant functions with matrix valued kernels.
\newblock {\em Journal of Machine Learning Research}, 8(Mar):385--408, 2007.

\bibitem{Rusnak2013}
Lucas~J. Rusnak.
\newblock Oriented hypergraphs: introduction and balance.
\newblock {\em Electron. J. Combin.}, 20(3):Paper 48, 29, 2013.

\bibitem{schafer2021sparse}
Florian Sch\"{a}fer, Matthias Katzfuss, and Houman Owhadi.
\newblock Sparse cholesky factorization by kullback--leibler minimization.
\newblock {\em SIAM Journal on Scientific Computing}, 43(3):A2019--A2046, 2021.

\bibitem{scholkopf2018learning}
Bernhard Scholkopf and Alexander~J Smola.
\newblock {\em Learning with kernels: support vector machines, regularization,
  optimization, and beyond}.
\newblock MIT Press, 2018.

\bibitem{srivastava2014dropout}
Nitish Srivastava, Geoffrey Hinton, Alex Krizhevsky, Ilya Sutskever, and Ruslan
  Salakhutdinov.
\newblock Dropout: a simple way to prevent neural networks from overfitting.
\newblock {\em The journal of machine learning research}, 15(1):1929--1958,
  2014.

\bibitem{tinhofer2012computational}
Gottfried Tinhofer, Rudolf Albrecht, Ernst Mayr, Hartmut Noltemeier, and
  Maciej~M Syslo.
\newblock {\em Computational graph theory}, volume~7.
\newblock Springer Science \& Business Media, 2012.

\bibitem{Wendland:2005}
H.~Wendland.
\newblock {\em Scattered data approximation}, volume~17 of {\em Cambridge
  Monographs on Applied and Computational Mathematics}.
\newblock Cambridge University Press, Cambridge, 2005.

\bibitem{williams1996gaussian}
Christopher K.~I. Williams and Carl~Edward Rasmussen.
\newblock {\em Gaussian Processes for Machine Learning}.
\newblock The MIT Press, 2006.

\bibitem{yoo2020deep}
Gene~Ryan Yoo and Houman Owhadi.
\newblock Deep regularization and direct training of the inner layers of neural
  networks with kernel flows.
\newblock {\em arXiv preprint arXiv:2002.08335}, 2020.

\bibitem{younes2010shapes}
Laurent Younes.
\newblock {\em Shapes and diffeomorphisms}, volume 171.
\newblock Springer, 2010.

\end{thebibliography}

\end{document}